%% file: thesis.tex
\documentclass[11pt]{report}         
\usepackage{utthesis2}
\usepackage[utf8]{inputenc}
\usepackage{graphicx}
\usepackage{graphics}
\usepackage{adjustbox}
\usepackage{amsmath}
\usepackage{algorithm}
\usepackage{algorithmic}
\usepackage{float}
\usepackage{booktabs}
\usepackage{placeins}
\usepackage{epsfig} 
\usepackage{amsmath} 
\usepackage{amssymb}  
\usepackage{tikz}
\usepackage{standalone}
\usepackage{bondgraphs}
\usepackage{url}
\usepackage{blox}
\usepackage{gensymb}
\usepackage{breakurl}
\usepackage{booktabs}
\usepackage{tabularx}
\usepackage{booktabs,amsfonts,dcolumn}
\usetikzlibrary{shapes,arrows}
\usetikzlibrary{calc, patterns, decorations.pathmorphing, decorations.markings, shapes.geometric, arrows, positioning, angles}
\DeclareMathOperator*{\argmax}{arg\,max}

\mastersthesis                     

\centerchapter                     

\oneandhalfspace                   

\renewcommand{\thesisauthor}{Henry Fielding Cappel}    

\renewcommand{\thesismonth}{December}     

\renewcommand{\thesisyear}{2020}      

\renewcommand{\thesistitle}{A Hierarchical Multi-Robot Mapping Architecture Subject to Communication Constraints}     

\renewcommand{\thesisauthorpreviousdegrees}{ }

\renewcommand{\thesissupervisor}{Luis Sentis}

\renewcommand{\thesiscommitteemembera}{Luis Sentis, Advisor}
\renewcommand{\thesiscommitteememberb}{Takashi Tanaka}

\renewcommand{\thesisauthoraddress}{hencappel1@gmail.com}

\renewcommand{\thesisdedication}{To my parents Jeff and Catherine}

\renewcommand{\thesisdegree}{MASTER OF SCIENCE IN ENGINEERING}  

\renewcommand{\thesisdegreeabbreviation}{MSE}

\begin{document}

\thesiscopyrightpage                 


\thesistitlepage                     

\thesissignaturepage                

\thesisdedicationpage                

\begin{thesisacknowledgments}        
I want to take the opportunity to thank everyone who helped me along the way towards completing my thesis and my master’s degree. Thank you to the University of Texas and specifically to Aerospace Engineering for providing me with an opportunity to explore my passion for robotics. 

A big thank you to my advisor Dr. Sentis for providing me with the opportunity to research deeply into the field of human centered robotics and supporting me with all my interests and giving me advice at just the right moments. Thank you to Dr. Tanaka for teaching me about Network Control Systems and supporting the development of my thesis.

Thank you to the Human Centered Robotics Laboratory and all members, coworkers, and most importantly friends that I have made along the way. A special thank you to Steven Jens Jorgensen for giving me the chance to do research at NASA Johnson Space Center and for having immense patience with me as I pestered him with questions. Thank you to Mihir Vedantam and Ryan Gupta for sharing my passion for robotics and providing great conversation. Thank you to Huang Huang, Binghan He, and Gray Thomas for collaborating with me to develop an improved exoskeleton and thank you Huang Huang for making six hours of experiments an enjoyable experience. Thank you to all the other members of the HCRL. I will thoroughly miss our thoughtful yet silly conversations and distractions. 

Thank you to all other friends and colleagues I have met along the way within aerospace engineering and the University of Texas as a whole. 

Finally, thank you to my parents Jeff and Catherine for supporting me, believing in me, and always reminding me that I am far more capable than I lead myself to believe.                                  
\end{thesisacknowledgments}          

\begin{thesisabstract}               
Multi-robot systems are an efficient method to explore and map an unknown environment. The simulataneous localization and mapping (SLAM) algorithm is common for single robot systems, however multiple robots can share respective map data in order to merge a larger global map. This thesis contributes to the multi-robot mapping problem by considering cases in which robots have communication range limitations. The architecture coordinates a team of robots and the central server to explore an unknown environment by exploiting a hierarchical choice structure. The coordination algorithms ensure that the hierarchy of robots choose frontier points that provide maximum information gain, while maintaining viable communication amongst themselves and the central computer through an ad-hoc relay network. In addition, the robots employ a backup choice algorithm in cases when no valid frontier points remain by arranging the communication relay network as a fireline back to the source.

This work contributes a scalable, efficient, and robust architecture towards hybrid multi-robot mapping systems that take into account communication range limitations. The architecture is tested in a simulation environment using various maps.
\end{thesisabstract}                 

\tableofcontents                     

\chapter{Introduction}              
\section{Area Background}
Intelligent robotic systems are not a new concept but have significantly advanced in recent years. The progression towards efficient network systems goes back to early computing days when operating systems relied on single process computing where one processor could only run a single process at a time. Later came the process of time sharing, which allowed multiple clients to access the capabilities of the central processor through a time slice such that the system could process multiple users' requests. Engineers improved further by developing client server architectures and object oriented programming as a way to increase efficiency and distribute computing over an extended network. 

In the context of robotics, a single robot is an inherent distributed system in the sense that each component of a robot plays a specific role as a piece of a larger system. Components can be classified as actuators, processors, sensors, mechanical components or a mixture of the four. In order to connect these components the robot has a communication network that links each of the components into a structure where the distribution of tasks are allocated such that the robot fulfills its overall objective. For example, a quadcopter may have an inertial measurement unit (IMU) and lidar sensor to take in data about its surrounding environment. The information is routed to a central processor that sends out commands to the actuators (in this case motors to spin the propellers). 

As an extension of a single robot system, multi agent systems, or more specifically multi robot systems, place the global objective of one robot as a sub task of a greater global system. Examples of these include drone swarms or networks of autonomous vehicles. An application of multi robot systems that has gained traction in recent years is multi robot mapping of unknown environments. In many instances a user may wish to deploy a team of robots to fulfill a task or set of tasks in an environment unknown to the robots or user. Developing an accurate map of the environment stands as the foundation of successful completion of other high level tasks and multi robot systems can provide accurate maps efficiently. 

It is no secret that efficiency lies in the proper distribution of tasks as can be seen in a well run workforce or organization, but how exactly to distribute operations from low level computing to high level robotic networks is an established question. 

\section{Research Question}
The research question of this thesis exists at the integration of three distinct areas of robotics; multi robot systems, simultaneous localization and mapping (SLAM), and robot exploration. Each of these subfields is well studied and has seen significant improvements in recent years. The question is how can one develop an efficient, robust, and scalable architecture for multi robot SLAM subject to communication constraints?

\section{Objectives and Challenges}
I will first discuss the objectives of this project in relation to answering the main research question and then explain the hurdles around achieving these goals. Before diving into the objectives of this project it should be noted that the system used for this research is a hybrid multi agent system. A hybrid system contains robot agents along with a central processing computer outlined in section 2.1.4. This type of system has advantages towards achieving the main objective of an efficient, scalable, and robust architecture for multi robot mapping.

The efficiency of the system will be determined by a minimal information exchange amongst agents. Information exchange, especially across WiFi networks, runs the risk of delay and drops. Systems that rely heavily on information exchange lose their functionality if communication fails at any point. Therefore, relying on minimal information exchange amongst agents and the central processor mitigates potential issues arising with communication channels. Efficiency is also determined by intelligent distribution of tasks by both the agents and central server. Tasks include information processing as well as exploration of the environment. The objective is to develop a system where each agent takes as much of the information processing as possible such that no one agent is overloaded with processing requirements and agents optimally distribute exploration tasks. Finally, the efficiency will be measured according to minimal time completion for developing a map of the environment. 

Scalability will be considered successful if the system can be extended to include many more agents without losing efficiency and robustness. An environment may be expansive and dense such that many agents are required to complete the mapping in a reasonable time. The objective of this research is to develop a system that can be used with small numbers of agents (2 or 3) or large numbers of agents (20 to 30).

Robustness will be measured according to how well the architecture can adapt to unforeseen challenges. These challenges may include robot mechanical or electrical failure, communication delays, communication drops, errors in the SLAM algorithm, unforeseen objects in the environment, and the unknown unknowns.

The final objective is to be able to map a range of environments. These environments may be indoor spaces full of objects and obstacles such as a warehouse, office building, or home, or more expansive outdoor spaces including college campuses, cities, or residential areas. The versatility of a multi robot mapping system expands a users capability to employ the system to a unique environment and prevents pigeonholing into specific structures. 

The main challenge with a multi robot system is communication range. Many robot systems rely upon WiFi or Bluetooth in order to transmit and receive data amongst agents. These platforms have strong data rates (up to 54 Mbps for WiFi and around 2 Mbps for Bluetooth), but are limited in range (100-150 feet for WiFi and around 33 feet for Bluetooth) \cite{hassan2012review}. This means that a system that relies upon either of these two platforms for inter-robot communication cannot permit the robots to travel beyond the available range. This is generally not problematic for indoor environments, but outdoor environments extending well beyond the WiFi range of any one robot may pose important issues.

The other challenge associated with communication is channel capacity for large sources of information. Although the channel capacities for both WiFi and Bluetooth are generally more than sufficient for certain data exchanges, there may be instances that robots want to send large amounts of data through the network such as video sharing where WiFi or Bluetooth modules are limited in capacity. In the scope of this project this does not pose to be an issue as map data is the largest amount of data transmitted amongst agents, but in the case of extending the architecture to a more information heavy context the data rates may become an issue.

The final challenge associated with a large multi robot system is collisions amongst agents. Many agents in a small, dense environment almost certainly restrict path planning and navigation. The coordination of agents to avoid collisions with each other and the environment while still completing the objective is an important consideration.

\section{Broader Impact}
\subsection{Space Applications}
Multi robot systems may prove to be the future of space exploration as they are resource efficient, relatively lightweight, more robust than single robot systems, and can perform dangerous yet necessary tasks that humans otherwise cannot perform. Space agencies are currently investigating robot systems for lunar and planetary tasks. The Japanese Aerospace Exploration Agency (JAXA) plans to use robots to construct lunar structures and setup a space based solar power system. Currently, the Mars Exploration Rovers (MER) launched by NASA in 2003 inhabit the Mars surface to investigate past water activity \cite{leitner2009multi}, however they operate independently of each other. Manned missions to Mars will likely be a reality in the next decades and multi robot systems will need to setup structures allowing humans to survive. As a foundation of building habitable structures, there is a need for a robust autonomous mapping architecture that allows the robot network to localize objects and other agents in the environment.

\begin{figure}[h]
    \centering{
    \includegraphics[width=\linewidth]{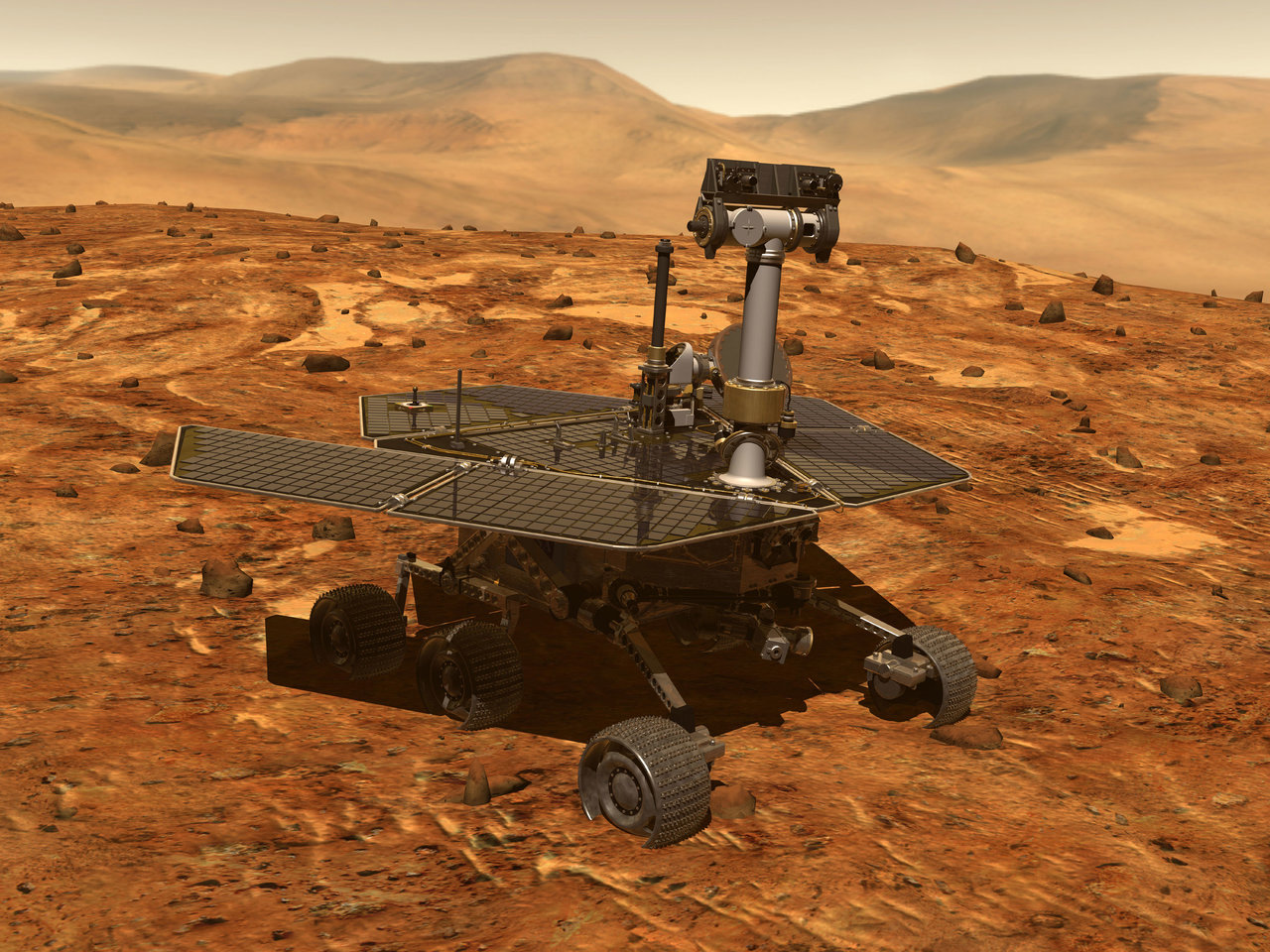}}
    \caption{The NASA mars exploration rovers (MER) Spirit and Opportunity may set the stage for a robot age of planetary exploration as well as space habitation construction. Image courtesy of \cite{mars_rover} }
    \label{fig:capbot_image}
\end{figure}

\subsection{Environmental Tasks}
Climate change and biodiversity losses call for a technological revolution that can help clean up bodies of water, mimic species playing a crucial role in the ecosystem, and monitor endangered species. Robots are advantageous in their ability to explore inaccessbile areas to humans and if used correctly provide non-invasive monitoring. \cite{van2018dawning} marks the distinction between two types of environmental robots; robots in ecology and ecological robots. The former classifies robots that collect ecological data used for research purposes. This may include a drone swarm deployed after a volcanic eruption to monitor air quality. The latter classifies robots that have a direct ecological role. These may include robots that mimic a certain species to fill a scarcity or target an invasive species to preserve an ecosystem. Multi robot systems provide a more efficient means to fulfill these tasks, especially if environments are expansive, by intelligently distributing tasks to more agents. Similar to space applications many environmental tasks may be in GPS denied environments forcing robot systems to develop accurate maps of the space.

\subsection{Domestic Tasks}
Robot integration into household chores has already taken hold. The ``Roomba" by iRobot traverses the floor of a house sweeping as it moves, but the ``Roomba" acts as an independent agent. Many common household chores can be outsourced to multi robot systems facilitating efficient completion of tasks. Multi robot systems may be able to assist with cleaning, surveillance, grocery delivery, or garbage collection \cite{cavallo2014development}. Additionally, an area that deserves more attention is multi robot systems for elderly care. Robots and specifically robot networks can provide an affordable option for in home care for elderly residents \cite{di2018multi}. Although robots do not provide the human to human interaction that is essential for caretaking, many of the mundane and tedious tasks of in home care can be given to the robots such that family members and caretakers are freed up for more personal interactions.  

\section{Ethical Considerations}
Certainly, the ethical considerations in a robotics age are not trivial. Multi robot systems provide an efficient and cost effective means to complete mundane and oftentimes repetitive tasks. However, a heavy reliance on these systems means that many human jobs can become obsolete leading to greater unemployment. It is important to weigh the advantages of increased productivity with the cost of taking jobs from people. Additionally, many of the resources used to establish robots in everyday life come at a cost to society and the environment \cite{van2018dawning}. This includes the waste produced when a robot reaches the end of its lifespan as well as the waste produced from energy consumption that is often not from renewable resources. The final ethical consideration to be considered is with human interpersonal connection. Human dependence on robot systems have the potential to deter human to human interaction if robots take the place of daily tasks. It is important to integrate robots into society such that human to human interaction is not heavily sacrificed.

\section{Thesis Roadmap}
This thesis will consist of the following main topics. Chapter 2 will provide the reader with detailed background information regarding multi agent systems, SLAM algorithms, robot path planning and navigation, an explanation of hierarchical structures and an extensive literature review of the aforementioned topics within the context of this thesis. The reader will find a detailed mathematical formulation of extended kalman filter (EKF) based SLAM as well as particle filter based SLAM extended to the use of occupancy grid maps. Additionally, as motivating examples, chapter 2 will also include an application of path planning implemented on the NASA Valkyrie humanoid robot and an application of a hierarchical information structure applied to whole body control that rely on similar algorithms presented in this project.

Chapter 3 is a detailed explanation of the methods used in this project including an overview of the project architecture, the algorithms developed for multi robot coordination, and specifications of the multi-robot system architecture including system type, the hierarchical structure, information exchanges amongst agents, and communication limits. The chapter will also include algorithms related to frontier point extraction, and robot choosing operations.

Chapter 4 provides the reader with simulation results based on two different maps, and a discussion of the potential benefits and contributions of this work.

In conclusion, chapter 5 will discuss how successfully the project has addressed the objectives outlined in section 1.3 and where there is need for further research. Finally, chapter 5 will provide possibilities for future work.

\chapter{Related Work} 
\section{Multi-agent Systems}
I will provide a brief introduction into multi agent systems providing definitions of relevant concepts within the context of this thesis. Additionally, I will provide a taxonomy of multi-robot systems explaining the variations in different structures.

\subsection{Definition of Agent}
The question, ``what defines an agent?" is the foundation for establishing the classifications of a multi-agent system. \cite{weiss1999multiagent} defines an agent as, ``autonomous, computational entities that can be viewed as perceiving their environment through sensors and acting upon their environment through effectors." Varying definitions exist for the concept of ``agent", but many overlap on core attributes such as autonomy, capacity to process and exchange information, follow causal structures, act towards objectives, and contain boundaries and states \cite{rocha2017introductory}. It is easy to assume an agent represents a physical robotic structure, but the term is far more general. For example an agent can be as physically defined as a drone within a drone swarm or as abstract as a function of code.

\subsection{Definition of Multi-agent System}
Multi-agent systems expand the definition of a single agent to a system containing multiple agents. \cite{stone2000multiagent} defines a multi-agent system as, ``a loosely coupled network of problem solving entities (agents) that work together to find answers to problems that are beyond the individual capabilities or knowledge of each entity (agent)." Once again, the idea of agent within this context remains general. The term ``multi-agent systems" represents an umbrella for many architectures including multi-robot systems described below.

\subsection{Definition of Multi-robot System}
Multi-robot systems are a subclass of multi-agent systems. They follow the same definitions as multi-agent systems with the added stipulation that each agent is a robot. The robots work together or in opposition in order to achieve some goal. The distinction of a multi-robot system is that interaction with the environment and other agents occurs through a physical medium from one physical structure to another.

\subsection{Taxonomy}
Multi-robot systems fall into a variety of distinct categories. The first to consider is organization. Three main organization structures classify multi-robot systems; centralized architectures, decentralized architectures, and hybrid architectures. A centralized architecture gives control authority to one agent to dictate the actions of other agents. In many instances the central agent takes on the majority of the computational burden and passes task commands to the other agents in the system. On the other hand, the second category, a decentralized architecture, distributes tasks amongst all the agents in the system. Decentralized systems are advantageous in computational efficiency as all agents are utilized to perform calculations, but limited in optimality due to incomplete information sharing. The third category, hybrid architectures, is a combination of the two. In many hybrid architectures a central agent will take on more computational burden and send commands to other agents, but also allocate tasks and decision making to other agents.

The second taxonomic category is robot composition. Multi-robot systems can be classified as either homogeneous (all robots are the same) or heterogeneous (robot models differ). Both categories come with costs and benefits. For example, homogeneous systems lack versatility. If a task or set of tasks requires many different capabilities homogeneous robots fall short. On the other hand homogeneous systems are easier to implement due to redundant hardware and system specifications. On the contrary, heterogeneous systems are more versatile, but require more tuning and considerations during implementation.

The third category is strategy type. Multi-robot systems can either be cooperative (robots working together to achieve a common goal) or competitive (robots working against one another for personal goals). An example of a cooperative system is two robots playing hide and seek trying to find a person in a house. The robots distribute tasks such that they both achieve the same global goal. On the other hand a competitive strategy may be two robots playing hide and seek against each other. One robot acts towards the goal of localizing the other robot, whereas the other robot works towards the goal of staying out of view of the first robot. Cooperative strategies can be broken down further into static and dynamic teams. A static team implies that information received by any agent is independent of its own or other robots' actions. A dynamic team implies the opposite \cite{yuksel2013stochastic}.

The fourth category is information structure, which is an important classification in terms of problem difficulty. There are two main types of information structures for a multi-robot system. The first is called classical information structure. This structure is generally easier to solve and can be reduced to standard optimal control/optimization problems. An example is a system with complete information sharing \cite{yuksel2013stochastic}. 
$$I_t^i = [y_{[1,t]},u_{[1,t-1]}]$$
where every agent $i$ receives full histories of measurements and control inputs from all other agents in the system. Non-classical information structures are generally more difficult. Examples of these include completely decentralized information structures \cite{yuksel2013stochastic}
$$I_t^i = [y_{[1,t]}^i]$$
where each agent only receives information of its own measurement history, or one-step delayed control sharing \cite{yuksel2013stochastic}
$$I_t^i = [y_{[1,t]}^i,u_{[1,t-1]}]$$
where each agent only has access to its own measurement histories and control input histories of all other agents. Quasi-classical or partially nested information structures lie between classical and non-classical information structures and many times produce tractable problems to solve. An example is one-step delayed information sharing \cite{yuksel2013stochastic}
$$I_t^i = [y_t^i,y_{[1,t-1]},u_{[1,t-1]}]$$

The fifth classification is communication. Communication can be divided into direct (transfer of information from one robot to another through on board hardware) or indirect (receive information from modifications in the environment) \cite{farinelli2004multirobot}. Direct methods of communication are more robust to information errors, but require a communication channel to transmit the data. On the other hand indirect methods require estimation techniques to gain information of other agents' states and are more prone to errors, but do not depend on communication channels. 

The sixth and final classification is system architecture. This dictates the strategy of a system to adapt to unforeseen changes. Deliberative architectures reorganize the system as a whole in response to unforeseen changes. This may include a central agent recalculating routes for all the agents in response to a changing environment. The other is a reactive architecture. Instead of a coordinated reorganization, each agent reactively changes its behavior to adapt to the change while still attempting to achieve its personal goal \cite{farinelli2004multirobot}.

\begin{table}[t]
\centering
\begin{adjustbox}{width=1\textwidth}
\begin{tabular}{|l|l|l|}
\hline
\textbf{Classification} & \textbf{Attribute}                 & \textbf{Definition}                                                                                                        \\ \hline
Organization            & Centralized                        & \begin{tabular}[c]{@{}l@{}}One central agent taking computational\\  load and giving commands to other agents\end{tabular} \\ \cline{2-3} 
                        & Decentralized                      & \begin{tabular}[c]{@{}l@{}}Tasks distributed amongst all agents.\\  Each agent has shared responsibility\end{tabular}      \\ \cline{2-3} 
                        & Hybrid                             & \begin{tabular}[c]{@{}l@{}}Combination of centralized \\ and decentralized\end{tabular}                                    \\ \hline
Robot Composition       & Homogeneous                        & One type of robot in the system                                                                                            \\ \cline{2-3} 
                        & Heterogeneous                      & \begin{tabular}[c]{@{}l@{}}More than one type of robot \\ in the system\end{tabular}                                       \\ \hline
Strategy                & Cooperative                        & \begin{tabular}[c]{@{}l@{}}Agents work together to achieve\\ a global goal\end{tabular}                                    \\ \cline{2-3} 
                        & Competitive                        & \begin{tabular}[c]{@{}l@{}}Agents work against one another \\ to achieve personal goals\end{tabular}                       \\ \hline
Information Structure   & Classical                          & \begin{tabular}[c]{@{}l@{}}Easy to solve\\ More information sharing\end{tabular}                                           \\ \cline{2-3} 
                        & Non-Classical                      & \begin{tabular}[c]{@{}l@{}}Difficult to solve\\ Less information sharing\end{tabular}                                      \\ \cline{2-3} 
                        & Partially Nested (Quasi-Classical) & \begin{tabular}[c]{@{}l@{}}Tractable problems\\ Sufficient information sharing\end{tabular}                                \\ \hline
Communication           & Direct                             & \begin{tabular}[c]{@{}l@{}}Communication amongst\\ agents via on board hardware\end{tabular}                               \\ \cline{2-3} 
                        & Indirect                           & \begin{tabular}[c]{@{}l@{}}Information gain via\\ environmental change\end{tabular}                                        \\ \hline
System Architecture     & Deliberative                       & \begin{tabular}[c]{@{}l@{}}Complete system adaptation\\ to unforeseen changes\end{tabular}                                 \\ \cline{2-3} 
                        & Reactive                           & \begin{tabular}[c]{@{}l@{}}Local robot adaptation to unforeseen \\ changes\end{tabular}                                    \\ \hline
\end{tabular}
\end{adjustbox}
\caption{Taxonomy of multi-robot systems}
\label{tabel 1}
\end{table}

\section{Simultaneous Localization and Mapping (SLAM)}
Simultaneous localization and mapping, or more commonly referred by its acronmym ``SLAM", is a hot topic for research within multi-robot systems. The algorithm builds a map of an unknown environment while also localizing the agent within the map. This parallel process is thought of as a chicken and egg type problem, because the parameters needed to build a map rely upon the robot's state (position and orientation), while the robot's state depends on the map generated. In some instances a user provides a pre-generated map to the robot such that the SLAM structure downgrades to a localization problem. However, oftentimes a map is unknown a-priori and the robot or team of robots need to build a map in real time. 

The SLAM algorithm works as an iterative likelihood estimate of both landmarks in the environment and the robot's state in relation to these estimated landmarks. Landmarks represent distinguishing features of the environment such as trees in a park, walls in a building, or other static objects in an area of space. These landmarks are determined based on sensor data of the robot itself. For instance a robot may use a Lidar rangefinder to collect range and bearing data of surrounding objects. The robot will estimate the landmarks' position with respect to it's own state, predict the state in the next step, and condition it's prediction on new sensor data received at the next timestep.

At its root the SLAM problem involves maximizing the likelihood of the probability distribution
$$p(m,x|u,z)$$
where $m$ is the state of all landmarks in the map, $x$ is the state of the robot, $u$ is the control input of the robot, and $z$ is the measurement of the robot through sensor data.

SLAM varies by technique and type of map generated. The two most common SLAM based algorithms, which will be outlined in great detail below, are based on the Extended Kalman Filter (EKF) and the particle filter. The SLAM algorithm can produce maps of varying type including volumetric (3D mappings), feature based maps (features or landmarks in an environment), topological maps, geometric maps (outline of a space), or occupancy grid maps (grid map indicating likelihood of obstacle at each grid point). An illustration of three of the maps is shown below.

\begin{figure}[h]
    \rule{0pt}{12pt}\\
    \centering{
    \resizebox{\columnwidth}{!}{
    \def\svgwidth{1.0\columnwidth}
    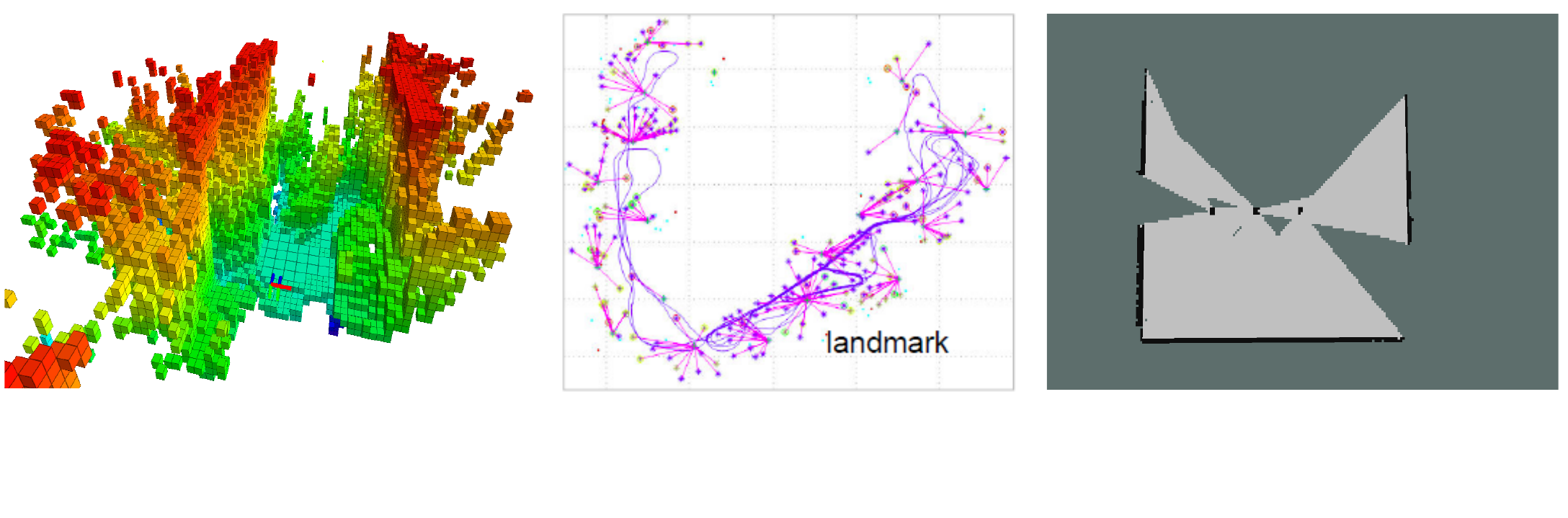}}
    \caption{An illustration of three types of maps generated by SLAM. a) represents a volumetric (3D) map, b) represents a landmark or feature based map, and c) represents an occupancy grid map. Images courtesy of \cite{volumetric_map_image}, \cite{feature_map_image}}
    \label{fig:map_types}
\end{figure}

\subsection{EKF based SLAM}
I will provide a detailed explanation of EKF based SLAM for estimating environmental landmarks and robot state. This derivation is based on the video lecture series \cite{stachniss_EKF_SLAM}.

The goal of EKF based SLAM is to estimate the state vector
$$ \mu = \begin{bmatrix}
    r_x  \\
    r_y \\
    r_{\theta} \\
    m_{1x} \\
    m_{1y} \\
    m_{2x} \\
    m_{2y} \\
    . \\
    . \\
    . \\
    m_{nx} \\
    m_{ny}
    \end{bmatrix} $$
where $r_x$, $r_y$, and $r_{\theta}$ are the robot state variables (position and orientation) and $m_{ix}$ and $m_{iy}$ are the $n$ landmark positions. Additionally, the covariance matrix can be represented as 
$$\Sigma = \begin{bmatrix}
    \Sigma_{r,r} && \Sigma_{r,m_1} && ... && \Sigma_{r,m_n} \\
    \Sigma_{m_1,r} && \Sigma_{m_1,m_1} && ... && . \\
    .&& && && .\\
    .&& && && .\\
    .&& && && .\\
    \Sigma_{m_n,r} && . && ... && \Sigma_{m_n,m_n} \\ 
\end{bmatrix}
$$
where subscript $r$ represents a condensed robot state and $m_i$ represents a condensed landmark state. The state estimate relies on the works of the Extended Kalman Filter. The EKF contains two stages; the prediction step and the update step. The prediction step uses a previous best state estimate and current control input to propagate this estimate based on the motion model of the system
$$\bar{\mu}_t = f(u_t,\mu_{t-1})$$
where $\bar{\mu}_t$ is the prediction estimate, $f$ represents the dynamics of the system, $u_t$ is the current timestep control input, and $\mu_{t-1}$ is the previous timestep state estimate. Similarly, it predicts the covariance matrix 
$$\bar{\Sigma}_t = F_t\Sigma_{t-1}F_t^T+R_t$$
where $\bar{\Sigma}_t$ is the predicted covariance matrix, $F_t$ is a linearized matrix of the motion model at the current timestep, and $R_t$ is the process noise of the system.

After determining the predicted state vector and covariance matrix, the EKF conditions these predictions based on the data received at the timestep. First, the kalman gain is computed as
$$K_t = \bar{\Sigma}_tH_t^T(H_t\bar{\Sigma}_tH_t^T+Q_t)^{-1}$$
where $K_t$ is the kalman gain, $H_t$ is the linearized matrix of the measurement model, and $Q_t$ is the measurement noise. Next, the state and covariance matrices are updated according to $$\mu_t = \bar{\mu}_t+K_t(z_t-h(\bar{\mu}_t))$$
$$\Sigma_t = (I-K_tH_t)\bar{\Sigma}_t$$
where $z_t$ is the measurement at timestep $t$. Algorithm 1 shows the full process.

\begin{algorithm}
\caption{Extended Kalman Filter Based SLAM}
\begin{algorithmic}
\REQUIRE $\mu_{t-1}$ //Previous state estimate
\REQUIRE $\Sigma_{t-1}$ //Previous covariance matrix

\STATE $\bar{\mu}_t = f(u_t,\mu_{t-1})$ //State prediction step
\STATE $\bar{\Sigma}_t = F_t\Sigma_{t-1}F_t^T+R_t$ //Covariance prediction step
\STATE $K_t = \bar{\Sigma}_tH_t^T(H_t\bar{\Sigma}_tH_t^T+Q_t)^{-1}$ //Kalman gain
\STATE $\mu_t = \bar{\mu}_t+K_t(z_t-h(\bar{\mu}_t))$ //State condition step
\STATE $\Sigma_t = (I-K_tH_t)\bar{\Sigma}_t$ //Covariance condition step
\STATE return $\mu_t$,$\Sigma_t$

\label{EKF}
\end{algorithmic}
\end{algorithm}

\subsection{Particle Filter Based SLAM}
Although EKF based SLAM is sufficient in many cases, it depends on assumptions of the system dynamics and probability densities. Namely, the EKF linearizes the system dynamics at each timestep, which in many cases does not pose a problem, but if system dynamics show strong nonlinearities the algorithm can deviate from truth. In addition, the EKF assumes an approximately gaussian posterior density. Dense environments can produce multi modal behavior of the posterior distribution and in these cases the EKF fails to accurately predict the state. 

The particle filter presents an improved strategy, but comes at a cost of increased computational load. In particular as the state space increases the computational efficiency dramatically decreases making the particle filter only viable for low-dimensional cases. I will outline the main mathematical features of the particle filter based on the lecture series \cite{stachniss_PF_SLAM}.

The particle filter uses a set of ``particles" along with corresponding weights as a sampling pool of possible states.
$$\chi = <x^{[j]},w^{[j]}>_{j=1,...,J}$$
The set $\chi$ contains $J$ particles of corresponding states ($x^{[j]}$) and weights ($w^{[j]}$). 

At each timestep the particle filter will initialize a new set $\chi_t$ as an empty set and use the previous timestep's set $\chi_{t-1}$ in order to resample according to each particles weight. After sampling from $\chi_{t-1}$ the particle filter propagates the sample according to the posterior
$$x_t^{[j]} \sim p(x_t|u_t,x_{t-1}^{[j]})$$
The propagation occurs with some uncertainty due to process noise, therefore the new particle is sampled according to the state posterior given control input $u_t$ and previous state estimate $x_{t-1}^{[j]}$.

The most important step of the particle filter is to calculate the weights of each newly sampled particle according to measurements received at the timestep. The weights are calculated according to 
$$w_t^{[j]} = p(z_t|x_t^{[j]})$$
Finally, the weights are normalized and added to the new set 
$$\chi_t = \chi_t + <x_t^{[j]},w_t^{[j]}>$$
which is used for the next sampling period. The full algorithm is presented in Algorithm 2.

In the context of SLAM the states of each particle may be the robot's position and orientation, and all positions of landmarks. However, in practice, this becomes computationally expensive as the number of particles required to cover the entire state space dramatically increases. As we will see in later sections the particle filter is used in combination with the EKF for a FastSLAM algorithm \cite{montemerlo2002fastslam}.

\begin{algorithm}
\caption{Particle Filter Based SLAM}
\begin{algorithmic}
\REQUIRE $\chi_{t-1}$ //Previous timestep set
\REQUIRE $u_t$ //Current timestep control input
\REQUIRE $z_t$ //Current timestep measurement

\STATE $\chi_t = \emptyset$ //Initialize set as empty
\FOR{$j=1$ to $J$} 
\STATE sample $x_t^{[j]} \sim p(x_t|u_t,x_{t-1}^{[j]})$ //Prediction step
\STATE $w_t^{[j]} = p(z_t|x_t^{[j]})$ //Update step
\STATE $\chi_t = \chi_t + <x_t^{[j]},w_T^{[j]}>$ //Append particle to set
\ENDFOR
\FOR{$j=1$ to $J$}
\STATE normalize $w_t^{[j]}$ //Normalize all weights in set
\ENDFOR
\STATE return $\chi_t$

\label{Particle_Filter}
\end{algorithmic}
\end{algorithm}

\subsection{Occupancy Grid Maps}
\begin{figure}[h]
    \centering{
    \includegraphics[width=0.5\textwidth]{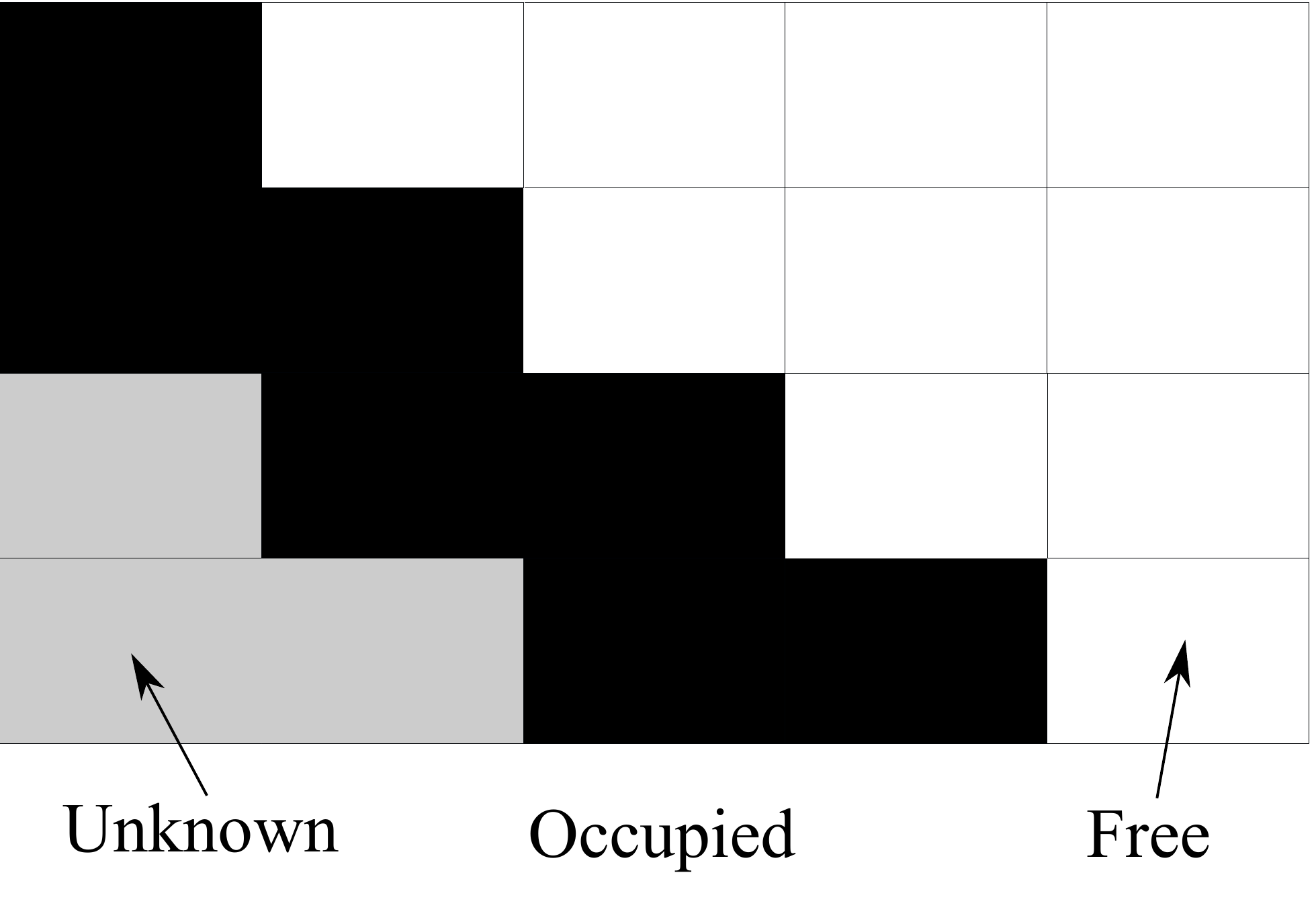}}
    \caption{An illustration of an occupancy grid map. The map is discretized into rectangular cells which are color coded corresponding to free (white), occupied (black), or unknown (grey)}
    \label{fig:occupancy_grid}
\end{figure}

Occupancy grid maps are commonly generated maps. I will provide a detailed explanation of their purpose. As their name suggests, occupancy grid maps are a discretized grid containing cells that represent obstacles, free space, or unknown space. An illustration of an occupancy grid map is shown in \ref{fig:occupancy_grid}.

Occupancy grid maps are advantageous because they do not require feature detectors. Unlike feature based maps, which depend upon an accurate feature detector to establish landmarks, every cell on the map receives a corresponding state. On the other hand occupancy grid maps are far more computationally expensive due to establishing every cell as a landmark \cite{stachniss_grid_SLAM}.

I will provide a full derivation of the algorithm to populate each of the grid map cells. This derivation is based on the video lecture \cite{stachniss_grid_SLAM}.

Two main assumptions establish a foundation for occupancy grid maps. First, each cell is assumed to be independent of all other cells, and, second, the world remains static. The probability density is expressed as
$$p(m|z_{1:t},x_{1:t})$$
where the map is conditioned on both the robot's measurements as well as the robot's state. It is important to note that this probability distribution assumes that the robot's state is known a-priori. Because each cell is independent of all other cells in the map we can write
$$p(m|z_{1:t},x_{1:t}) = \prod_i p(m_i|z_{1:t},x_{1:t})$$
According to Bayes' rule this can be rearranged as
$$p(m_i|z_{1:t},x_{1:t}) = \frac{p(z_t|m_i,z_{1:t-1},x_{1:t})p(m_i|z_{1:t-1},x_{1:t})}{p(z_t|z_{1:t-1},x_{1:t})}$$
Due to a markov property of measurements and knowledge of the state of the cell this can be rewritten as 
$$p(m_i|z_{1:t},x_{1:t}) = \frac{p(z_t|m_i,x_t)p(m_i|z_{1:t-1},x_{1:t-1})}{p(z_t|z_{1:t-1},x_{1:t})}$$
We can use Bayes' rule on the term $p(z_t|m_i,x_t)$ to write
$$p(z_t|m_i,x_t) = \frac{p(m_i|z_t,x_t)p(z_t|x_t)}{p(m_i|x_t)}$$
which leads to the full probability distribution
$$p(m_i|z_{1:t},x_{1:t}) = \frac{p(m_i|z_t,x_t)p(z_t|x_t)p(m_i|z_{1:t-1},x_{1:t-1})}{p(m_i)p(z_t|z_{1:t-1},x_{1:t})}$$
It is important to note that the term $p(m_i|x_t)$ is reduced to $p(m_i)$ due to the grid map structure. If we take the ratio of the probability that a cell is occupied with the probability that a cell is unoccupied we arrive at the expression
$$\frac{p(m_i|z_{1:t},x_{1:t})}{1-p(m_i|z_{1:t},x_{1:t})} = \frac{p(m_i|z_t,x_t)p(m_i|z_{1:t-1},x_{1:t-1})(1-p(m_i))}{(1-p(m_i|z_t,x_t))(1-p(m_i|z_{1:t-1},x_{1:t-1}))p(m_i)}$$
This is the key step in that it allows one to reclaim the original probability distribution in the form
$$p(m_i|z_{1:t},x_{1:t}) = (1 + \frac{(1-p(m_i|z_t,x_t))}{p(m_i|z_t,x_t)}\frac{(1-p(m_i|z_{1:t-1},x_{1:t-1}))}{p(m_i|z_{1:t-1},x_{1:t-1})}\frac{p(m_i)}{(1-p(m_i))})^{-1}$$
If we look closely we can see that three separate useful terms appear in the expression.
$$\frac{(1-p(m_i|z_t,x_t))}{p(m_i|z_t,x_t)}$$
represents the inverse sensor model of the robot,
$$\frac{(1-p(m_i|z_{1:t-1},x_{1:t-1}))}{p(m_i|z_{1:t-1},x_{1:t-1})}$$
represents a recursive term from previous states and sensor observations, and
$$\frac{p(m_i)}{(1-p(m_i))}$$
is a prior term of the state of the cell. This allows us to form a general algorithm for updating each of the individual cells in the occupancy grid as the robot receives more information from the environment.




\subsection{FastSLAM}
FastSLAM is an efficient SLAM algorithm that combines the advantages of the EKF and particle filter into a single algorithm \cite{montemerlo2002fastslam}. EKF based SLAM and particle filter based SLAM alone pose significant drawbacks. EKF based SLAM cannot account for highly nonlinear system dynamics as well as highly non-gaussian distributions. On the other hand particle filter based SLAM produces a computational infeasibility for high state space dimensions. Therefore, FastSLAM combines the two based on a Rao-Blackwellization of the posterior distribution
$$p(x_{0:t},m_{1:M}|z_{1:t},u_{1:t}) = p(x_{0:t}|z_{1:t},u_{1:t})p(m_{1:M}|x_{0:t},z_{1:t})$$
which can be rewritten as
$$p(x_{0:t},m_{1:M}|z_{1:t},u_{1:t}) = p(x_{0:t}|z_{1:t},u_{1:t})\prod_{i=1}^Mp(m_i|x_{0:t},z_{1:t})$$
by accounting for the independence of each landmark. In this form the posterior density can be broken into two parts. The first piece $p(x_{0:t}|z_{1:t},u_{1:t})$ can be estimated with a particle filter to determine robot poses. The second piece $\prod_{i=1}^Mp(m_i|x_{0:t},z_{1:t})$ can be estimated with $M$ 2x2 EKFs. The SLAM problem is now broken into two components; standard pose estimation and mapping with known poses \cite{stachniss_FastSLAM}.

The adaptation for occupancy grids is fairly straightforward. Instead of M 2x2 EKF's for map landmark updates, the algorithm uses Algorithm 3 to update each grid cell. In addition, the algorithm uses scan matching to correct robot poses by taking into account previous map information and current sensor information.
$$x_t^* = \argmax_{x_t}p(z_t|x_t,m_{t-1})p(x_t|u_{t-1},x_{t-1}^*)$$
and in some cases the algorithm uses an improved proposal distribution for choosing particles based on current sensor information \cite{stachniss_FastSLAMGrid}.
$$x_t^{[k]} \sim p(x_t|x_{1:t-1}^{[k]},u_{1:t},z_{1:t})$$

In conclusion, FastSLAM is an efficient SLAM algorithm that uses Rao-Blackwellization in order to divide the posterior density such that the EKF and particle filter work together to produce accurate and efficient map estimates.

\subsection{Map Merging}
Map merging is a key component of multi-robot SLAM, because multiple robots can share local map information presented as occupancy grid maps and overlay the maps to produce a larger global map. This is especially useful if robots are spread out in distant locations, because a global map presents each robot with much more information of the environment than each local map. 

Map merging algorithms have two main challenges. The first challenge is feature detection amongst the maps. This process allows the map merging algorithm to determine the features of the map that are significant for overlay with other maps. In theory robots that map the same area should produce maps that have a significant number of features in common. Two approaches used for feature detection include Scale-Invariant Feature Transform (SIFT) and Speeded Up Robust Features (SURF). However, both of these are patented and not available for open source use and both have limitations. Another algorithm is called Oriented FAST and Rotated BRIEF (ORB), which is open source and ameliorates the issues of SIFT and SURF \cite{horner2016map}.

The second challenge is to estimate the transformation amongst individual maps. \cite{horner2016map} uses an algorithm based on image stitching for panoramic images to estimate the affine transformation amongst individual maps and extends this to a global map transformation. The algorithm uses random sample consensus (RANSAC) to estimate the transformation, which can produce more accurate results by focusing on inliers rather than fitting to all data (inliers and outliers).

The advantage of the algorithm presented in \cite{horner2016map} is that initial robot poses do not need to be known. Oftentimes it is difficult for robots to directly estimate the pose of other robots in the system, therefore the algorithm is useful by estimating affine transformations based on map overlaps.

\begin{figure}[h!]
    \rule{0pt}{8pt}\\
    \centering{
    \resizebox{\columnwidth}{!}{
    \def\svgwidth{1.0\columnwidth}
    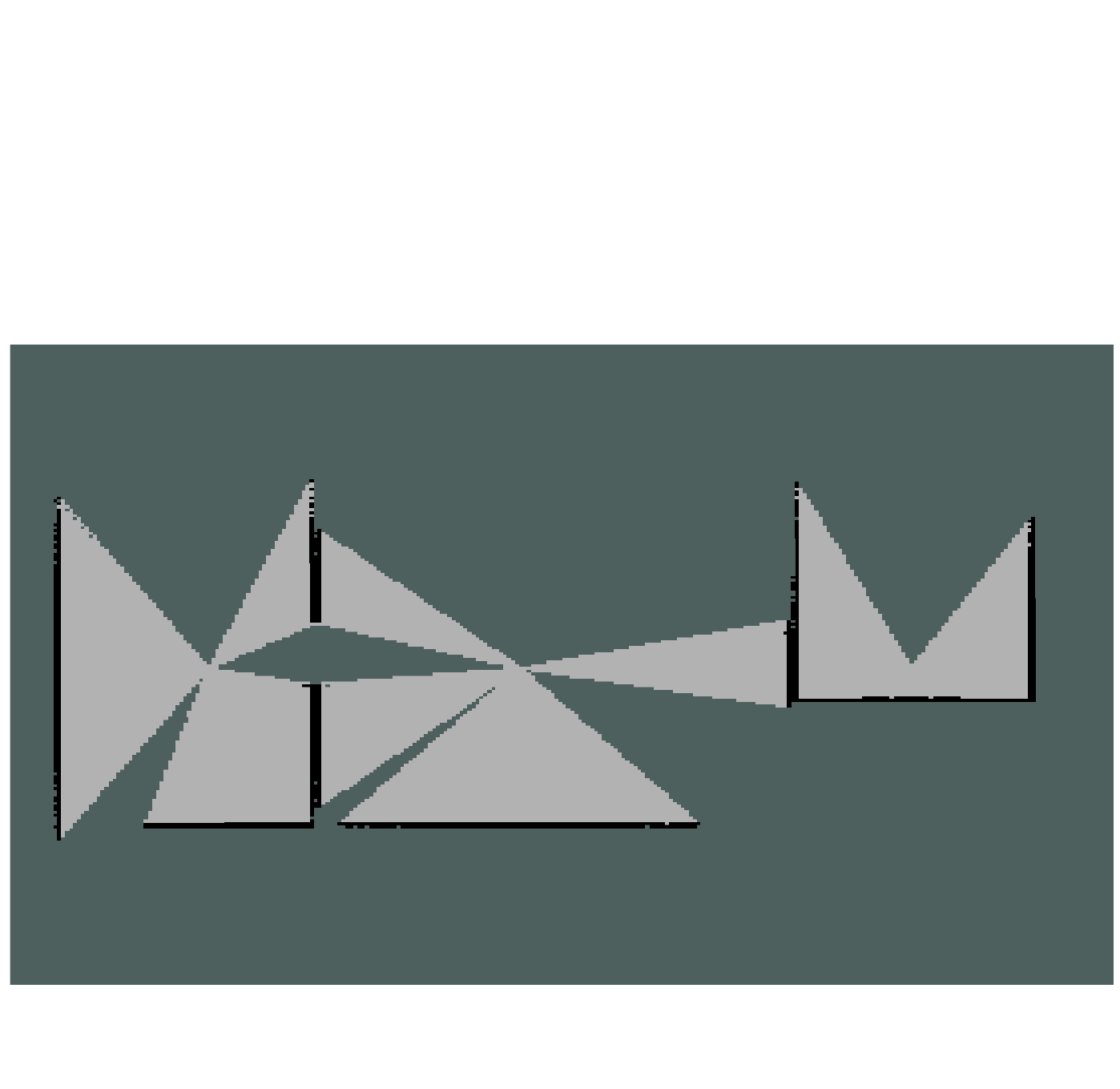}}
    \caption{An illustration of the map merging algorithm presented in \cite{horner2016map}. Each of the local maps (a),(b),(c) are stitched together through the ROS navigation multirobot\_map\_merge package. The resulting map is shown in (d)}
    \label{fig:map_merge}
\end{figure}

The algorithm presented in \cite{horner2016map} is provided in the multirobot\_map\_merge package included with the ROS navigation stack. The package is used in this project and will be explained in detail in later sections. A visualization of the result of the map merging algorithm is shown in figure \ref{fig:map_merge}.


\section{Path Planning and Navigation}
Path planning is crucial for robots to safely traverse a space from point A to point B. Without path planning and navigation, one could not guarantee a mobile robot network to function safely and effectively. The navigation procedure is generally divided into a global planning algorithm to produce a safe path around static and known obstacles and a local plan to adhere to the global plan while avoiding dynamic and unknown obstacles. A common global planning algorithm, A*, is outlined below and an application towards a humanoid robot is included as an illustrating example.

\subsection{A* Global Path Planner}
The A* graph traversal algorithm is based on Dijkstra's algorithm for finding shortest path in a graph, but uses informed search heuristics to achieve better results. Nodes are explored based on a priority queue determined by a heuristic function
$$f(n) = g(n) + h(n)$$
where $f(n)$ represents the total edge cost of the node $n$, $g(n)$ is the cost of traversing the edge from the start node to the current node, and $h(n)$ is the edge cost of the current node to the goal node. 

At each iteration the algorithm chooses the node in the priority queue with lowest heuristic cost. Neighbor nodes are explored with the chosen node as the parent. The algorithm determines the heuristic cost of each neighbor, whether or not the node has already been explored, and updates the priority queue accordingly. The algorithm is repeated until the node within a tolerance of the goal is chosen at which a path is reconstructed. The complete A* algorithm is shown in algorithm 4.

In some instances a weighted A* is preferred to reduce time duration at the cost of optimality. The weighted A* aggressively chooses nodes that lead towards the goal in order to open less total nodes in the graph \cite{jorgensen2020finding}. The heuristic function becomes
$$f(n) = g(n) + \epsilon h(n)$$
where $\epsilon > 1$ represents a weight placed on the edge cost from current node to goal.

\begin{algorithm}
\caption{A* Search Algorithm}
\begin{algorithmic}
\REQUIRE StartPose
\REQUIRE GoalPose

\STATE ENQUEUE(OpenSet,StartPose)
\STATE StartPose $\rightarrow$ Gscore := 0
\STATE StartPose $\rightarrow$ Fscore := 0

\WHILE{OpenSet is not empty}
\STATE current := node in OpenSet with lowest Fscore
\IF{current within tolerance of GoalPose}
\STATE reconstruct path
\ENDIF
\FOR{each neighbor of current}
\STATE TentativeGscore := current $\rightarrow$ Gscore + d(current,neighbor)
\IF{TentativeGscore $<$ neighbor $\rightarrow$ Gscore}
\STATE neighbor $\rightarrow$ parent := current
\STATE neighbor $\rightarrow$ Gscore := TentativeGscore
\STATE neighbor $\rightarrow$ Fscore := neighbor $\rightarrow$ Gscore + d(neighbor,GoalPose)
\IF{neighbor not in OpenSet}
\STATE ENQUEUE(OpenSet,neighbor)
\ENDIF
\ENDIF
\ENDFOR
\ENDWHILE
\STATE return empty path

\label{algorithm_a_star}
\end{algorithmic}
\end{algorithm}

\subsection{Path Planning for Humanoid Robots}

\begin{figure}[h!]
    \centering{
    \includegraphics[width=1.0\textwidth]{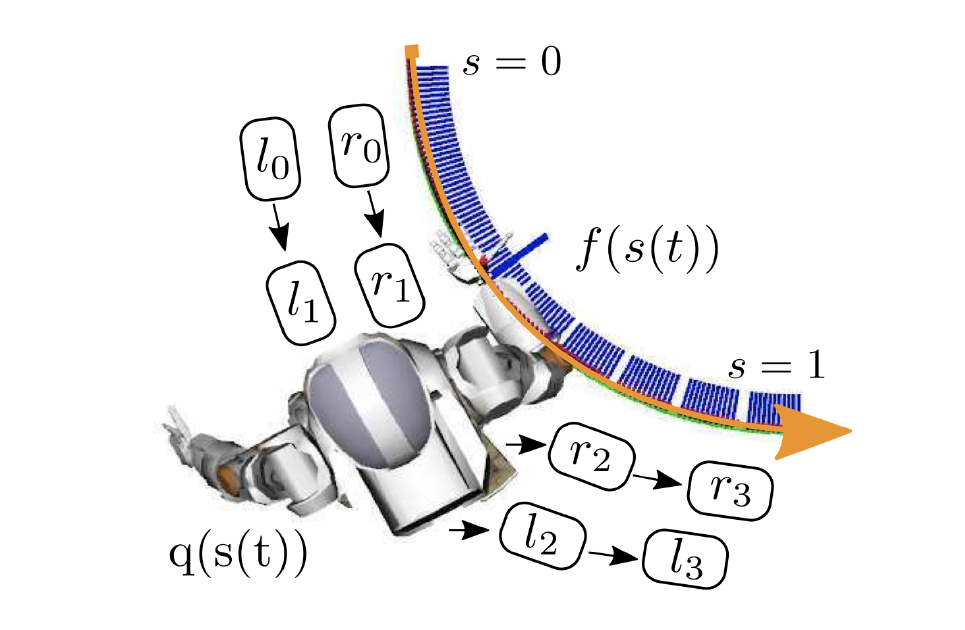}}
    \caption{A locomanipulation planner based on a weighted A* search algorithm implemented on the NASA Valkyrie humanoid robot. The planner generates plans based on locomotion progression, manipulation progression, or locomanipulation progression. Footsteps are denoted by $l_i$ and $r_i$ and the end effector progression is denoted by $s$. Image courtesy of \cite{jorgensen2020finding}}
    \label{fig:valkyrie}
\end{figure}

We can apply the weighted A* path planning algorithm to find locomanipulation trajectories for the NASA Valkyrie humanoid robot \cite{jorgensen2020finding}. Locomanipulation is a difficult problem for humanoid robots, because locomotion (moving around an environment) and manipulation (moving objects within an environment) often destabilize the robot if done in parallel. As a result most locomotion and manipulation tasks are performed sequentially. However, the goal of this work was to generate feasible locomanipulation plans quickly such that the robot's end-effector reaches the goal pose.

In order to generate valid locomanipulation trajectories it is useful to fix a locomotion plan and find feasible manipulation trajectories based on this locomotion constrained manifold. In essence, the trajectory of the end-effector is calculated only within the nullspace of the locomotion plan. 

We use a weighted A* search algorithm to generate plans based on this model, where edge transitions can be one of three options: pure locomotion, pure manipulation, or locomanipulation. The edge cost function becomes
$$\Delta g(v_1,v_2) = w_s (1-s) +w_{step} + w_{L} r(v_2)$$
where $\Delta g(v_1,v_2)$ is the cost of transition, $w_s (1-s)$ is the cost associated with moving the end effector according to the $s$ variable, $w_{step}$ is the cost associated with taking a footstep, and $w_L r(v_2)$ penalizes states that deviate from a particular body path. The important thing to consider is that edge transitions are now a function of two variables (footstep poses, and end effector poses ($s$)). This allows the planner to consider transitions that progress both locomotion and manipulation trajectories simultaneously.

In order to speed up the process of determining feasible plans a neural network classifier is used to prune nodes that produce infeasible trajectories. The classifier's output is considered in the cost function 
$$\Delta g(v_1,v_2) = w_s (1-s) +w_{step} + w_{L} r(v_2) + w_d(1-n(v_1,v_2))$$
where $w_d(1-n(v_1,v_2))$ is a cost associated with feasibility of edge transition determined by the neural network. This ensures that the planner chooses nodes that adhere to feasibility while still progressing towards the goal.

\begin{figure}[h!]
    \centering{
    \includegraphics[width=1.0\textwidth]{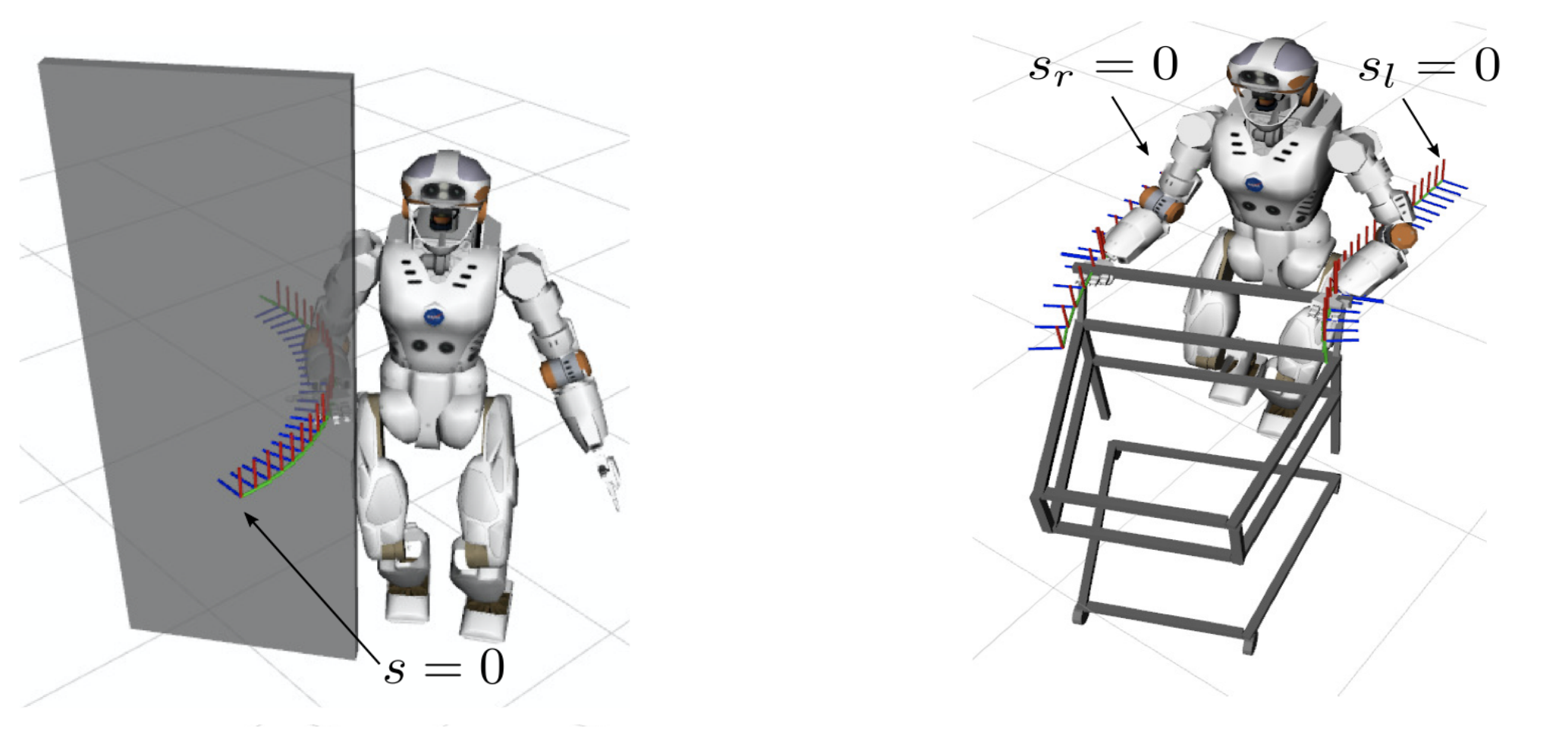}}
    \caption{The locomanipulation formulation is tested using two tasks. The left task involves moving the end effector from $s=0$ to $s=1$ in order to open a door. The right task involves progressing left and right end effectors ($s_l$ and $s_r$ respectively) towards pushing a cart. Images courtesy of \cite{jorgensen2020finding}}
    \label{fig:valkyrie_tasks}
\end{figure}

As a final improvement we use a weighted A* cost heuristic. The edge cost function becomes
$$\Delta g(v_1,v_2) = w_h(w_s (1-s)) +w_{step} + w_{L} r(v_2) + w_d(1-n(v_1,v_2))$$
in order to quickly progress the plans towards completing the end effector task. This formulation was validated on a cart pushing task and a door opening task as shown in figure \ref{fig:valkyrie_tasks}.

\section{Hierarchical Structures}
Hierarchies are a common approach towards solving multi agent problems. For example they can be used in the context of multi-robot task allocation \cite{hawley2013hierarchical}, multi-robot reinforcement learning \cite{cai2013combined}, multi-robot mapping \cite{chand2013mapping}, and even whole body control \cite{sentis2007synthesis}. Hierarchical structures are advantageous in many scenarios, because they eliminate the need to solve a global optimal solution involving all agents, but instead solve optimal solutions within a hierarchy.

The premise of a hierarchical structure is that agents are ranked according to a criteria. In the case of multi-robot systems the robots may be ranked according to capabilities. In homogeneous cases ranking may be arbitrary as all agents are classified in similar ways. Solutions are computed according to the priority queue of the hierarchy. For instance, the first agent may compute a solution and all following agents will compute solutions according to the null space of preceding agents.

This structure is especially common in the context of whole body control. Robots contain $n$ degrees of freedom, which means that joints can often interfere with other joints if the robot is expected to perform multiple simultaneous tasks. A hierarchy is setup in order to prevent this issue as tasks are prioritized according to importance and any task is performed only within the nullspace of preceding tasks.

\subsection{Hierarchical Structures Within Whole Body Control}
\begin{figure}[h!]
    \rule{0pt}{8pt}\\
    \centering{
    \resizebox{\columnwidth}{!}{
    \def\svgwidth{1.0\columnwidth}
    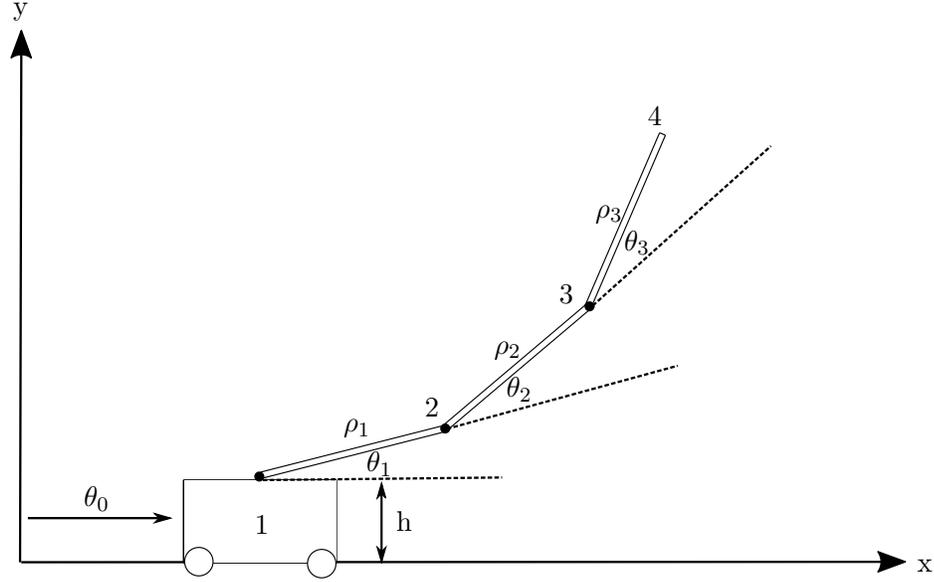}}
    \caption{The kinematic diagram for this robot. Each of the end effectors are labelled as 1,2,3,4 respectively.}
    \label{fig:kindiagram}
\end{figure}

As previously mentioned, whole body control uses a hierarchical task space representation to ensure that no specified tasks interfere. Tasks can be end effector position or velocity in cartesian space. I will provide an example of a whole body control structure within the kinematics domain using a hierarchical task space, and compare two optimal control strategies for solving the problem. The example will be based on the robot shown in figure \ref{fig:kindiagram}.

\subsubsection{Task Space Representation}
I will give a brief overview of how the joint space of a whole body control structure can be mapped into a task space using a series of null space projections. The mapping from a joint displacement to the end effector displacement in cartesian space is described by
\begin{equation} \label{eq2.1}
    dx_i = J_idq
\end{equation}
where $dx_i$ is the end-effector displacement of the ith task, $J_i$ is the jacobian of the ith task, and $dq$ is the joint displacement. 

If we define 
\begin{equation} \label{eq2.2}
    dq = dq_1+N_1dq_2+N_1N_{2/1}dq_3+...+N_{[s]}dq_s
\end{equation}
where
\begin{equation} \label{eq2.3}
    N_{[s]} = \prod_{i=1}^{s} N_{i/i-1}
\end{equation}
$N_{i/i-1}$ is the null space projector defined to be
\begin{equation} \label{eq2.4}
    N_{i/i-1} = I - (J_iN_{i-1/i-2})^{\dagger}(J_iN_{i-1/i-2})
\end{equation}
where
\begin{equation} \label{eq2.5}
    N_{1/0} = N_{1} = (I - J_1^{\dagger}J_1)
\end{equation}

By plugging \ref{eq2.2} into \ref{eq2.1} for every i in the task space we can establish a set of decoupled task space equations of the form
$$dx_1 = J_1dq_1$$
$$dx_2 = J_2dq_1+J_2N_1dq_2$$
$$dx_3 = J_3dq_1+J_3N_1dq_2+J_3N_1N_{2/1}dq_3$$

With this set of task space equations we can generate a discrete time state space model of the form
$$\begin{bmatrix}
x_{t+1}^{1} \\
x_{t+1}^{2} \\
x_{t+1}^{3} \\
. \\
. \\
. \\
x_{t+1}^{k}
\end{bmatrix}
=
I^{k \times k} \begin{bmatrix}
x_t^{1} \\
x_t^{2} \\
x_t^{3} \\
. \\
. \\
. \\
x_t^{k}
\end{bmatrix}
+
\begin{bmatrix}
J_1 && 0 && 0 && ... && 0 \\
J_2 && J_2N_1 && 0 && ... && 0 \\
J_3 && J_3N_1 && J_3N_1N_{2/1} && ... && 0 \\
. \\
. \\
. \\
J_k && ... && ... && ... && J_k\prod_{i=1}^{k-1}N_{i/i-1}
\end{bmatrix}
\begin{bmatrix}
dq_t^{1} \\
dq_t^{2} \\
dq_t^{3} \\
. \\
. \\
. \\
dq_t^{k}
\end{bmatrix}
+
\begin{bmatrix}
w_t^{1} \\
w_t^{2} \\
w_t^{3} \\
. \\
. \\
. \\
w_t^{k}
\end{bmatrix}
$$

\subsubsection{Task Hierarchy}

\begin{figure}[h]
    \rule{0pt}{12pt}\\
    \centering{
    \resizebox{\columnwidth}{!}{
    \def\svgwidth{0.57\columnwidth}
    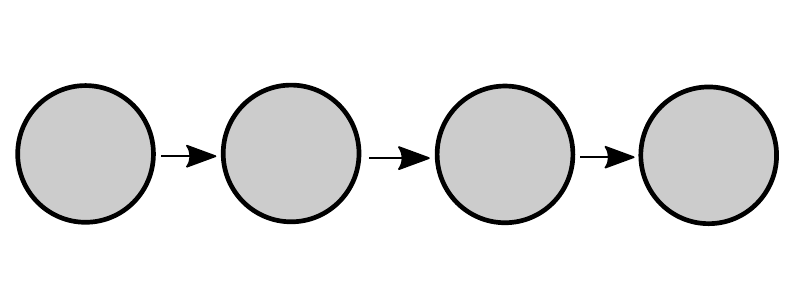}}
    \caption{A representation of the information structure for this system. Task 1 represents the prioritized task and all information flows downstream to lower level tasks.}
    \label{fig:hierarchy}
\end{figure}

Due to the sequential null space projections of tasks it is easy to establish a hierarchy of task objectives defined as end-effector position in cartesian space as shown by the discrete time state space model in the previous section. The top prioritized task generates a control policy based on information from it's own state, whereas all lower level tasks rely on information from their own state as well as control inputs of all higher level tasks. The information structure is displayed in \ref{fig:hierarchy} as well as equations \ref{eq3.1},\ref{eq3.2},\ref{eq3.3},\ref{eq3.4}.

\begin{equation}\label{eq3.1}
    u_1(t) = d\tilde{q}_t^{1} = \gamma_{1,t}(x_t^{1})
\end{equation}
\begin{equation}\label{eq3.2}
    u_2(t) = d\tilde{q}_t^{2} = \gamma_{2,t}(x_t^{2},u_t^{1})
\end{equation}
\begin{equation}\label{eq3.3}
    u_3(t) = d\tilde{q}_t^{3} = \gamma_{3,t}(x_t^{3},u_t^{1},u_t^{2})
\end{equation}
\begin{equation}\label{eq3.4}
    u_k(t) = d\tilde{q}_t^{4} =  \gamma_{k,t}(x_t^{k},u_t^{1},u_t^{2},u_t^{3},...,u_t^{k-1})
\end{equation}

\subsubsection{Optimal Control Formulation}
The aim of the control policies is to minimize the cost function
\begin{equation}
    J = \limsup\limits_{T\rightarrow \infty}\frac{1}{T}\sum_{t=1}^{T}\mathbf{E}(\tilde{x}_t^{T}Q\tilde{x}_t+u_t^{T}Ru_t)
\end{equation}
Because the information structure contains a hierarchy we can solve this system with the sparsity dynamic programming approach from \cite{lamperski2015optimal}. There is no time delay in this system therefore the solution reduces to $k$ solutions to the discrete backward Ricatti recursion. Due to this system containing a time varying B matrix the standard LQR dynamic programming approach solves a feedback gain matrix K at each timestep by solving the backward Ricatti equation.
$$P_{t-1} = Q+A^TP_tA-A^TP_tB(R+B^TP_tB)^{-1}B^TP_tA$$
$$P_T = I$$
$$K_t = (R+B^TP_{t+1}B)^{-1}B^TP_{t+1}A$$

However if we consider a sparsity structure the feedback gain $K_t^{s}$ is solved for each node where $s$ corresponds to the information node. The control input is reconstructed according to a propagation of the noise multiplied by each K matrix. If we let $$ \zeta_{t+1}^s = \sum_{w^{i}\rightarrow s}I^{s,{i}}w_t^i$$ where $$\zeta_0^s =  \sum_{w^i\rightarrow s}I^{s,i}x_0^i$$ and let $$x_0^i \sim \mathcal{N}(0,\sigma^2)$$ then the control input can be reconstructed as $$u_t = \sum_s I^{{i},s}K_t^s\zeta_t^s$$
This formulation ensures a sparsity on K and significantly decreases computation for the control input.

The weighting matrices $Q$ and $R$ are chosen as the identity matrices in order to put equal weight on control input and state error. These are initial choices for these parameters and may be updated in the future work.

\subsubsection{Results and Figures}
\begin{figure}[htp]
    \centering
    \includegraphics[width=1.0\textwidth]{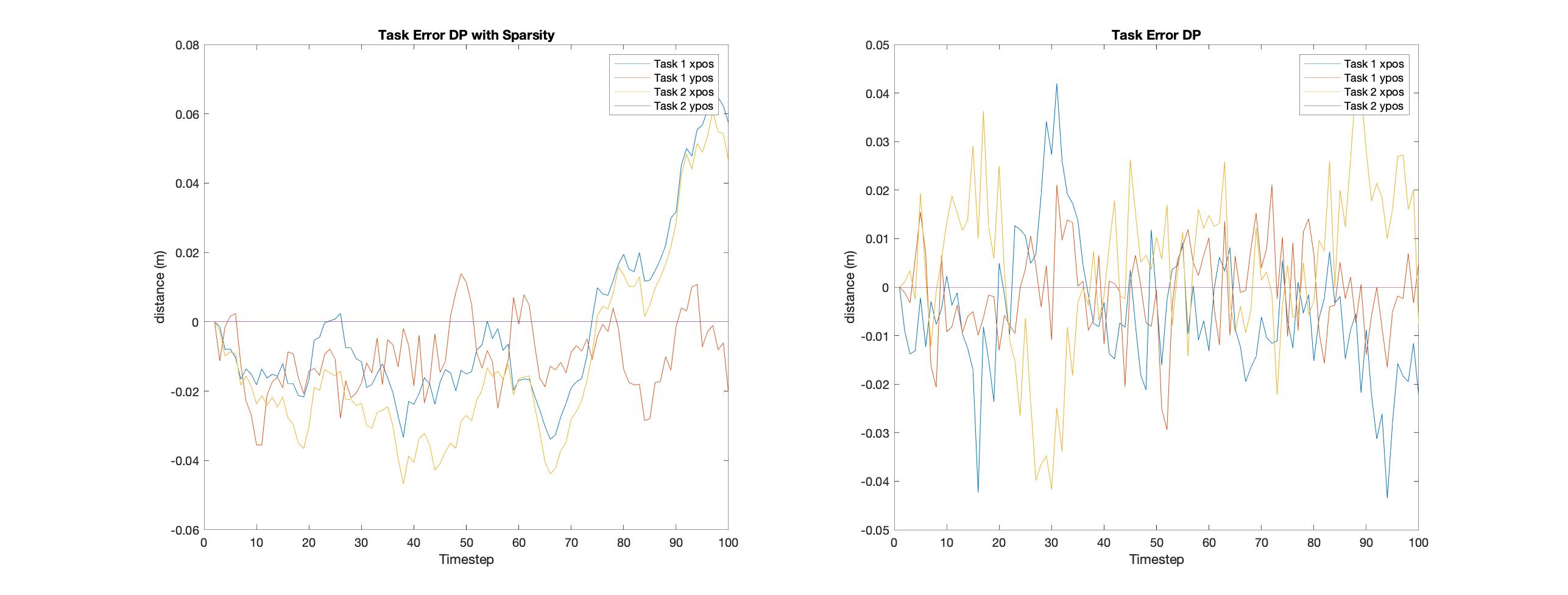}
    \caption{Task errors over time for dynamic programming and dynamic programming with sparsity}
    \label{fig:Error}
\end{figure}

\begin{figure}[htp]
    \centering
    \includegraphics[width=1.0\textwidth]{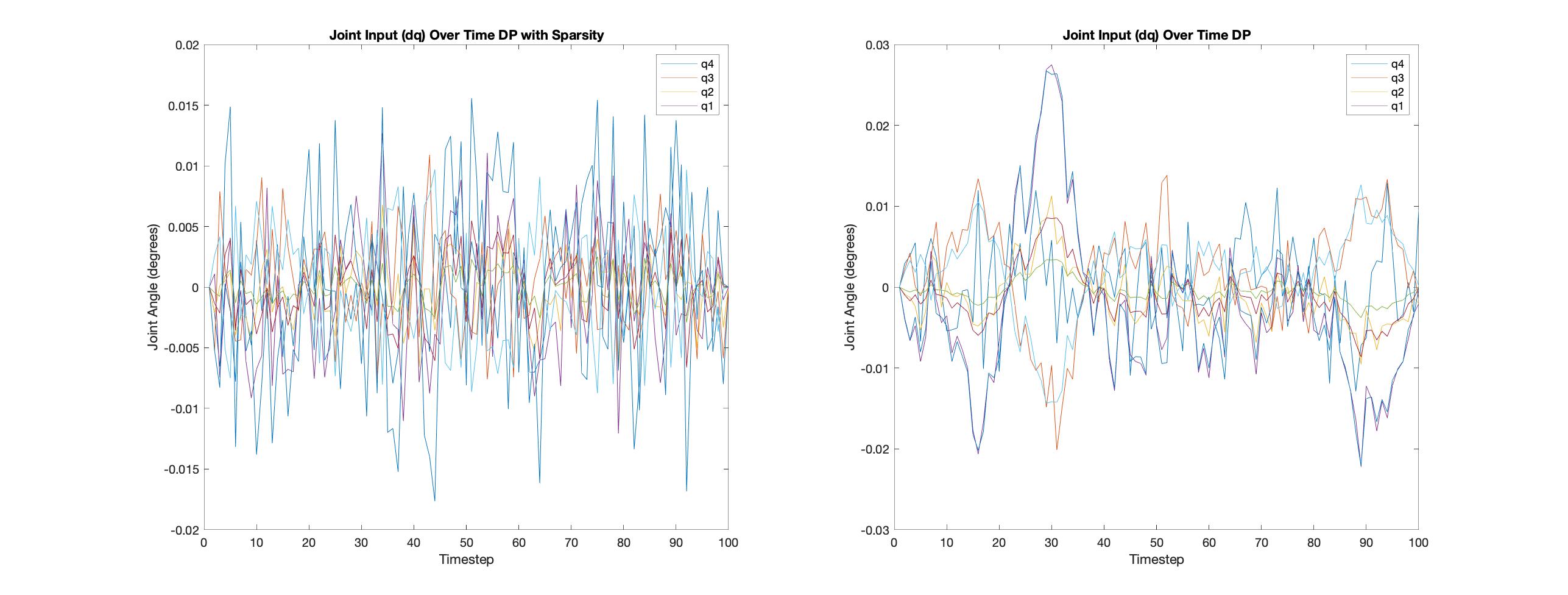}
    \caption{Joint inputs over time for dynamic programming and dynamic programming with sparsity}
    \label{fig:Input}
\end{figure}

\begin{figure}[htp]
    \centering
    \includegraphics[width=1.0\textwidth]{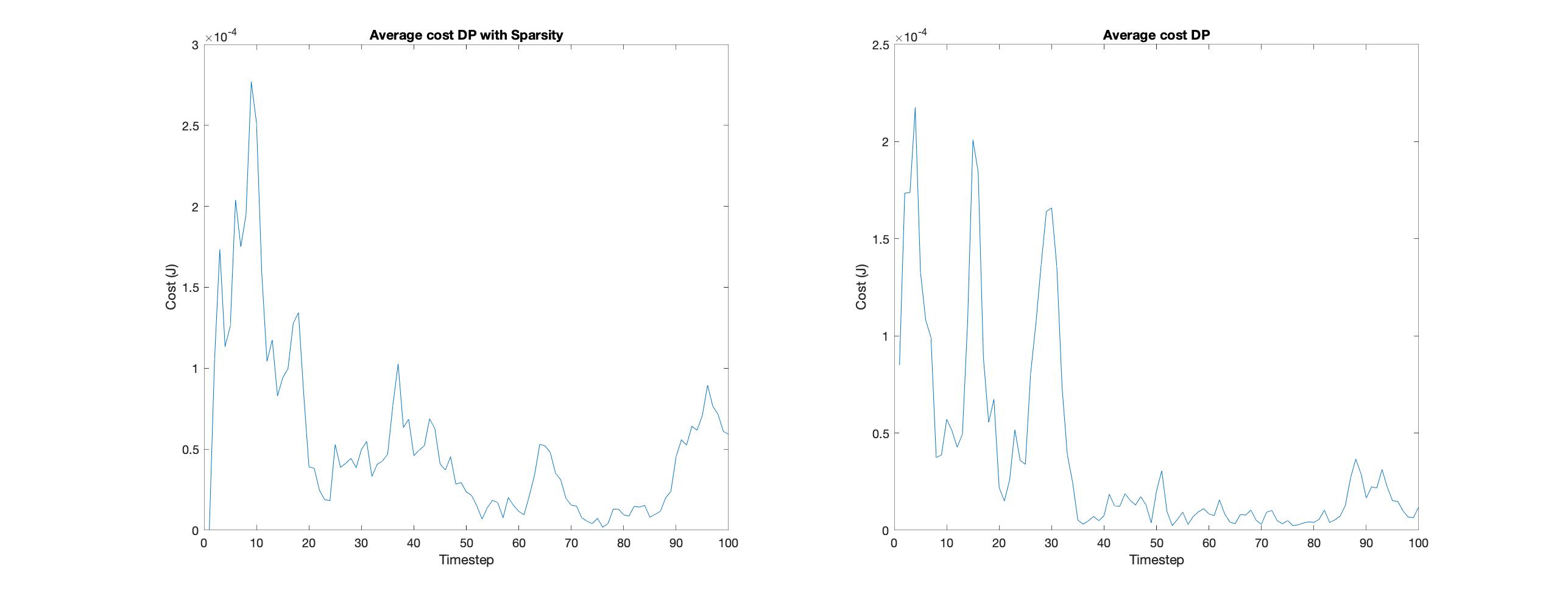}
    \caption{Average cost over time for dynamic programming and dynamic programming with sparsity}
    \label{fig:Cost}
\end{figure}

The results of a simulation comparing the control performances between standard dynamic programming using the backward Ricatti recursion and dynamic programming with sparsity is shown below. Figure \ref{fig:ActualTasks} shows the actual joint trajectories overlayed on nominal joint trajectories for both approaches. Figure \ref{fig:Error} shows the error of both approaches as a function of time. Figure \ref{fig:Input} shows the input commands for both approaches as a function of time, and figure \ref{fig:Cost} shows the average cost of both approaches as a function of time.

\begin{figure}[htp]
    \centering
    \includegraphics[width=1.0\textwidth]{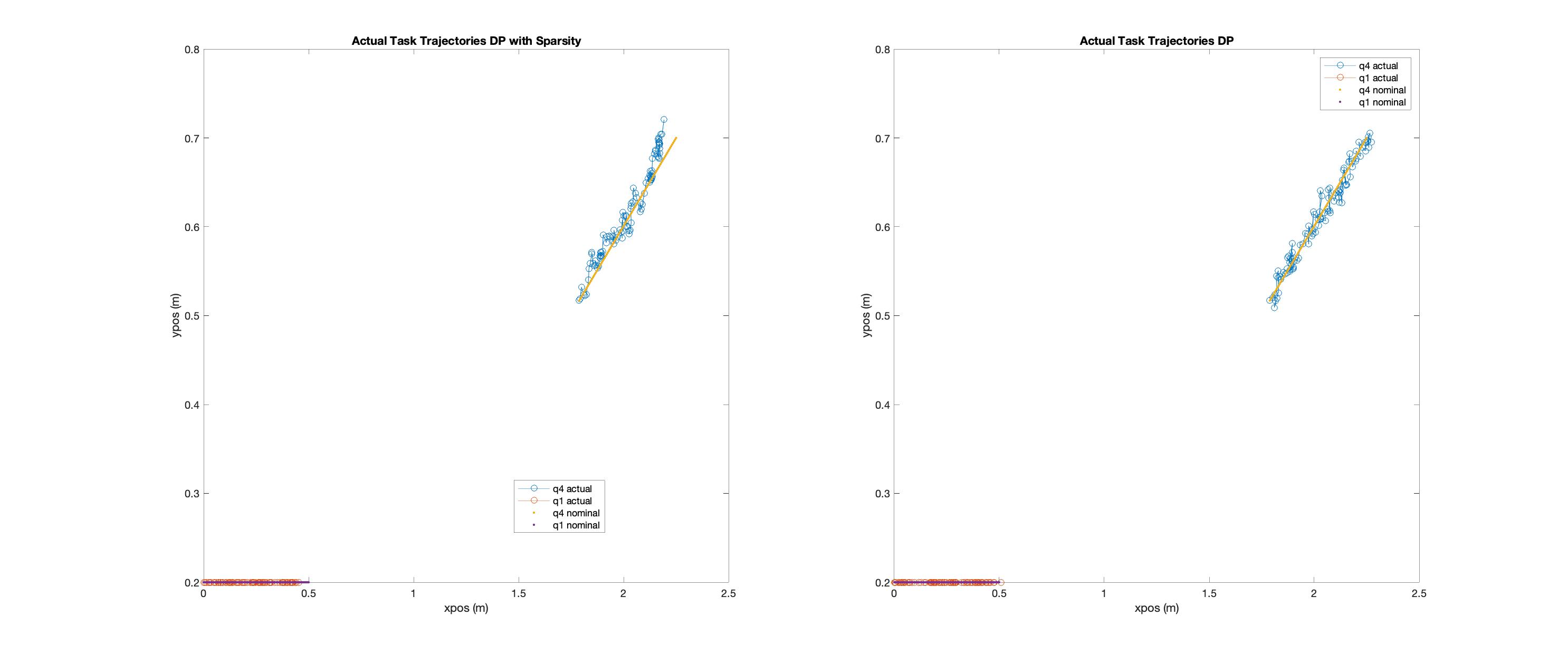}
    \caption{Actual simulated task trajectories for both dynamic programming and dynamic programming with sparsity}
    \label{fig:ActualTasks}
\end{figure}

\newpage
\section{Literature Review}
The multi-robot SLAM problem is not a novel research area. However, researchers have proposed a number of strategies to coordinate robots, merge local maps, and execute the SLAM algorithm. 

The types of maps generated and measurement sensors vary. For example, \cite{deutsch2016framework} generates graph based maps using a vision camera, \cite{choudhary2016multi} uses RGBD cameras to generate landmark based maps, and \cite{ozkucur2009cooperative} uses a laser rangefinder to create occupancy grid maps. 

Some researchers present novel methods to augment the local maps of individual agents, but do not consider communication limitations. \cite{gil2010multi} uses stereo cameras to perform visual based SLAM estimating landmarks with a FastSLAM algorithm. The research is based on a hybrid multi-robot system as local robots use FastSLAM to generate local maps and send these maps to a central server to perform map merging. However, the researchers do not address communication issues and assume perfect communication with the central server. In addition, \cite{ozkucur2009cooperative} compares the difference between using a standard EKF and FastSLAM for building a global landmark based map for a multi-robot system. Robots share map and state information on a local level and do not rely on a central server for map merging, but robot communication is not guaranteed as robots only communicate as they come within range of one another. 

On the other hand \cite{deutsch2016framework} and \cite{choudhary2016multi} present novel methods to augment the maps of individual agents and consider communication limitations. \cite{deutsch2016framework} combines local robot pose graphs into a global graph based map using visual markers and uses a correction strategy to handle communication delays amongst the robots. \cite{choudhary2016multi} compares 3D object based maps to pointcloud based maps for a decentralized multi-robot system. The researchers take advantage of local robot computational capacity by allocating object detection and object pose estimation to individual agents in order to reduce communication burdens. 

In addition, \cite{atanasov2015decentralized} considers computational efficiency and a hierarchical structure to develop control policies for a multi-robot SLAM system. The researchers develop control policies by minimizing the entropy of the system conditioned on future measurements. The strategy exploits sparsity in the planning process and reduces computational burden. The formulation considers a partially nested information structure and a hierarchy of sensors to develop a hierarchical control policy.

\cite{rooker2007multi}, \cite{banfi2018strategies}, \cite{de2009role}, and \cite{cesare2015multi} consider multi-robot exploration tasks in which robots are subject to, specifically, communication range limitations. \cite{amigoni2017multirobot} draws the distinction among three classes of communication range constraints; continuous, event-based, and none. \cite{banfi2018strategies} develops a strategy for event-based communication (robots only communicate when novel, relevant information is acquired). The researchers use a central computer to determine robot poses based on a centralized process and then extend the framework to a distributed process where robots choose their goal poses. \cite{rooker2007multi} considers continuous communication range restrictions. The researchers use a population sampling method to determine robot control policies based on a heuristic function taking into account viable distance amongst robots and the central source. Although agents maintain viable network structures, the coordination is completely centralized and can run into computational burdens for the central computer. 

\cite{de2009role} and \cite{cesare2015multi} solve the communication range issue with ad-hoc information relay networks. Robots use one another to relay information either amongst themselves or back to the source. This gives the robot network a means to explore areas beyond the range of any one robot.

Many researchers implement frontier based methods to explore the unknown of a map, but it is worth noting that other methods provide efficient results. \cite{liu2016leveraging} uses an artificial potential field to guide a system of robots. The potential field is based on attractors (unexplored areas of the map) and repulsors (other robots, obstacles, and already explored areas). Additionally, \cite{andries2015multi} and \cite{kuyucu2015superadditive} propose multi-robot exploration methods based on ant colonies, where robots leave traces to signal already explored areas to other robots.

\chapter{Methods}
\section{Project Overview}

\begin{figure}[h!]
    \centering{
    \includegraphics[width=0.87\textwidth]{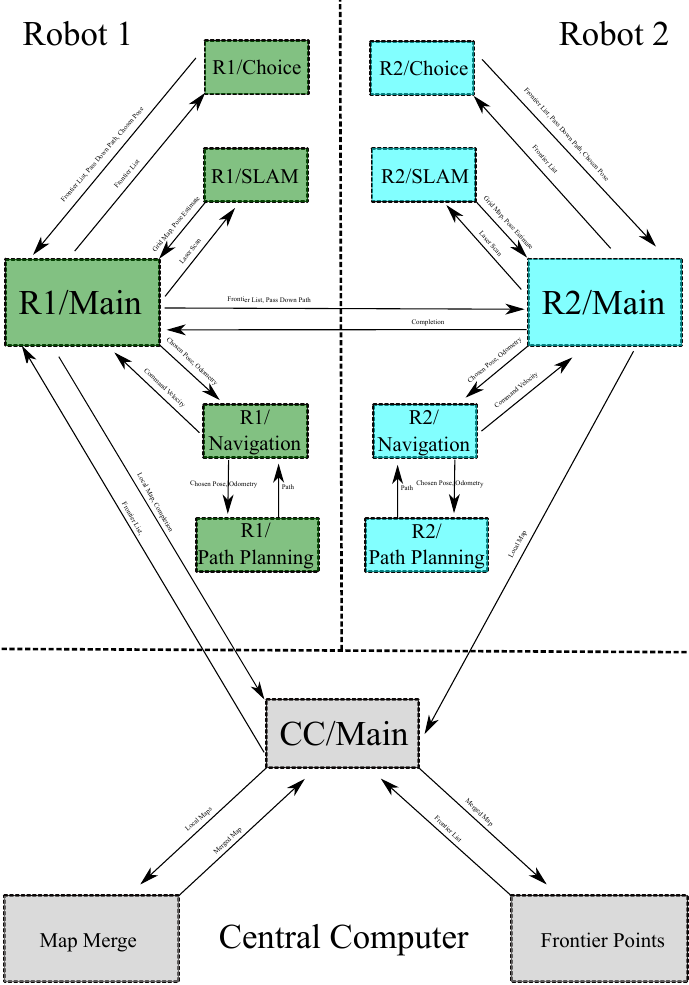}}
    \caption{The full system architecture. Every rectangle represents a node within each agent (separated by dashed lines). The arrows between nodes represent information exchanges with message types written above the lines }
    \label{fig:full_architecture}
\end{figure}

The objective of this project is to develop and test an autonomous multi-robot mapping architecture that is constrained by robot communication limits. The goals are to adhere to three main criteria; efficiency, scalability, and robustness. 

The project combines multi-robot systems, SLAM, and robot exploration into a cohesive architecture. Multiple robots communicate with the central computer while autonomously exploring and gathering information for building the map of the space. The robots are constrained by communication range limits that prevent exploration of distant points outside of a viable communication network. However, the robots can form an ad-hoc relay network to pass information from any one robot to another robot eventually reaching the central computer. The architecture attempts to optimize the coordination of the robot system such that the agents explore as much of the map as possible, while remaining in network communication amongst themselves and the central computer.

The robots arrange themselves in an ad-hoc communication relay network by exploiting a hierarchical structure explained in section 3.4.2. 

The full system architecture is shown in diagram \ref{fig:full_architecture}. Each box corresponds to a node within each robot or the central computer. Each node serves a particular function and the network of nodes and information sharing constitutes the entire architecture. Up until this point I have explained the nodes related to SLAM, navigation/path planning, and map merging. In the following sections I will provide a detailed explanation of the frontier point exploration and the robot coordination algorithms. Finally, I will provide an overview of the system architecture including the classification of the multi-robot system, hierarchical structure, information exchange, and communication constraints.

\section{Frontier Exploration}
\subsection{Frontier Point Extraction}
The central computer determines frontier points from the merged map in order to send goal poses to the hierarchy of robots. Frontier points are the threshold between the known and unknown of the map. In other words the frontiers indicate the areas that still need to be explored. The algorithm for finding these points is presented in Algorithm 5.

The algorithm is based on the Wavefront Frontier Detector algorithm presented in \cite{topiwala2018frontier} and modified to prune irrelevant frontiers and extract optimal frontier points. The search algorithm uses a double breadth first search in order to search the map. The outer breadth first search searches the map as a whole looking for relevant frontier points, while the inner breadth first search searches the span of an individual frontier. The algorithm ensures that no point occurs within more than one frontier. This prevents frontier points from being double counted on the final frontier list. 

Following frontier determination, frontiers are sorted to ensure that irrelevant frontiers either from mapping errors or spaces too small for robots to search are not appended to the frontier list. The algorithm sorts based on a size threshold of each frontier array. Only arrays above this threshold are considered. 

\begin{figure}[H]
    \centering{
    \includegraphics[width=1.0\textwidth]{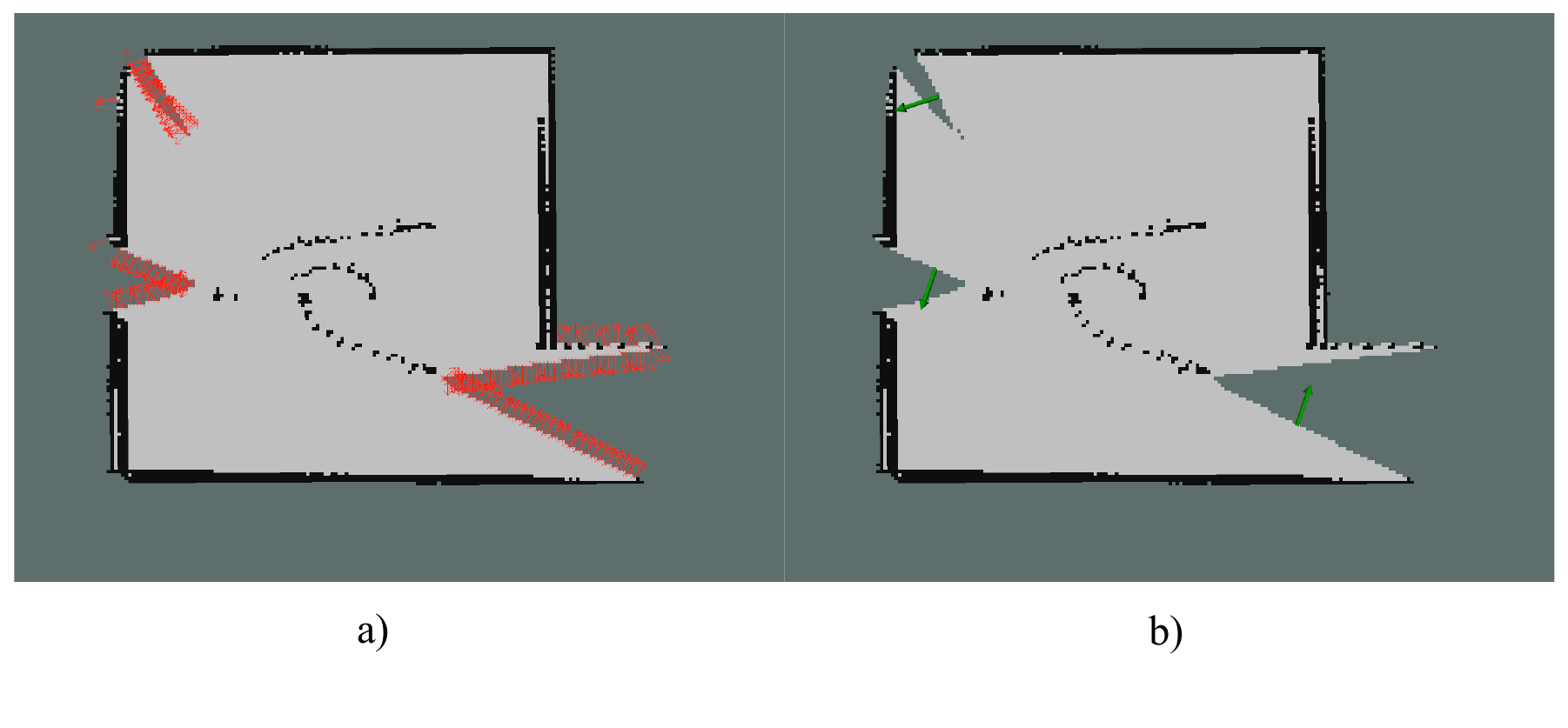}}
    \caption{An illustration of the frontier point search algorithm. a) shows all frontier arrays (red) and b) shows only the representative frontier points (green) of each frontier array}
    \label{fig:frontier_pts}
\end{figure}

Finally, the frontier point that is the spacial average of each relevant frontier array is chosen as the representative point for the frontier and appended to the frontier list. An illustration of the complete set of frontier points and the final frontier list is shown in \ref{fig:frontier_pts}.

\subsection{Frontier Hierarchy}
The algorithm sorts the final frontier list according to respective frontier array size. This assumes that frontier arrays with the largest number of points indicate areas of the map that have the most potential information gain. The hierarchical frontier list is then used for the robot coordination algorithms explained in section 3.3.

\begin{algorithm}
{\fontsize{11pt}{11pt}\selectfont
\caption{Frontier Points Search Algorithm}
\begin{algorithmic}
\REQUIRE MapQueue
\REQUIRE FrontierQueue 
\REQUIRE Pose

\STATE MapQueue $\leftarrow \emptyset$
\STATE ENQUEUE(MapQueue,Pose)
\STATE Pose $\rightarrow$ MapOpen = true
\WHILE{MapQueue not empty}
\STATE $current_m$ $\leftarrow$ DEQUEUE(MapQueue)
\IF{$current_m$ $\rightarrow$ MapClosed == true}
\STATE continue
\ENDIF
\IF{$current_m$ is a frontier point}
\STATE FrontierQueue $\leftarrow \emptyset$
\STATE NewFrontier $\leftarrow \emptyset$
\STATE ENQUEUE(FrontierQueue,$current_m$)
\STATE $current_m$ $\rightarrow$ FrontierOpen = true
\WHILE{FrontierQueue not empty}
\STATE $current_f$ $\leftarrow$ DEQUEUE(FrontierQueue)
\IF{$current_f$ $\rightarrow$ MapClosed, FrontierClosed == true} 
\STATE continue
\ENDIF
\IF{$current_f$ is a frontier point}
\STATE append $current_f$ to NewFrontier
\FORALL{neighbors of $current_f$}
\IF{$neighbor_f$ $\rightarrow$ FrontierOpen, FrontierClosed, MapClosed != true}
\STATE ENQUEUE(FrontierQueue,$neighbor_f$)
\STATE $neighbor_f$ $\rightarrow$ FrontierOpen = true
\ENDIF
\ENDFOR
\ENDIF
\STATE $current_f \rightarrow$ FrontierClosed = true 
\ENDWHILE
\IF{size of NewFrontier is greater than threshold}
\STATE OptimalFrontierPoint = average pose of NewFrontier
\STATE append OptimalFrontierPoint to FrontierList
\ENDIF
\ENDIF
\FORALL{neighbors of $current_m$}
\IF{$neighbor_m \rightarrow$ MapOpen,MapClosed != true and $neighbor_m$ has at least one open space neighbor}
\STATE ENQUEUE(MapQueue,$neighbor_m$)
\STATE $neighbor_m \rightarrow$ MapOpen = true
\ENDIF
\ENDFOR
\STATE $current_m \rightarrow$ MapClosed = true
\ENDWHILE
\STATE return FrontierList

\label{algorithm_frontier_pts}
\end{algorithmic}
}
\end{algorithm}

\section{Robot Coordination}
Robot coordination is the backbone of a successful architecture. Robots choose frontier points as navigation goals based on the frontier list in section 3.2 such that each robot locally optimizes it's choice. I will outline the data structures used for the choice algorithm, the main algorithm itself, and a backup choice algorithm in cases when no valid frontier points remain. 

\subsection{Data Structures}
The choice algorithm depends on three main data structures. The first is the frontier list received from the preceding robot. The frontier list contains all available frontier points arranged according to the hierarchy of section 3.2.2. The second is the chosen pose that each robot sends to the navigation/path planning algorithm in order to reach the pose. The third and final structure is the pass down path that marks pose dependencies for receding robots such that the current robot can navigate to the chosen pose and use other robots as relays to maintain valid information exchange with the system. 

\subsection{Choice Algorithm}
The choice algorithm guarantees that a robot either chooses the best available frontier list pose or adheres to a dependency of a preceding robot for communication relay. The full algorithm is shown in algorithm 6. I will explain each of the main features of the algorithm in detail below.

\begin{algorithm}
\caption{Choose Frontier Point}
\begin{algorithmic}
\REQUIRE FrontierList
\REQUIRE PassDownPathList 

\IF{PassDownPathList not empty}
\STATE ChosenPose $\leftarrow$ DEQUEUE(PassDownPathList)
\STATE erase ChosenPose from FrontierList
\STATE return ChosenPose,FrontierList,PassDownPathList
\ELSE
\FORALL{FrontierPts in FrontierList}
\IF{FrontierPt is valid}
\STATE ENQUEUE(ValidFrontierList,FrontierPt)
\STATE find neighbors of FrontierPt
\ENDIF
\ENDFOR
\WHILE{ValidFrontierQueue not empty}
\STATE CandidatePose $\leftarrow$ DEQUEUE(ValidFrontierQueue)
\STATE compute valid paths of CandidatePose
\STATE PathList := CandidatePose $\rightarrow$ ValidPaths
\IF{PathList not empty}
\STATE ChosenPose := CandidatePose
\STATE PassDownPathList := OptimalPath
\STATE return ChosenPose,PassDownPathList,FrontierList
\ENDIF
\ENDWHILE
\IF{fail to find ChosenPose}
\STATE perform BackupChoiceAlgorithm
\ENDIF
\ENDIF

\label{algorithm_choose_main}
\end{algorithmic}
\end{algorithm}

The algorithm begins by checking a dependency from preceding robots in the hierarchy. The PassDownPathList is a list of frontier points that a preceding robot asks of the current robot to inhabit. It is important to note that each of the poses in the PassDownPathList are other frontier points determined from algorithm 5. If the dependency list is not empty then the robot automatically chooses the first pose in the list and erases this pose from the FrontierList.

If the PassDownPathList presents an empty array then the robot continues a normal choosing operation based on the FrontierList. The robot first filters the list by checking to see if each point is within a valid number of relay points to the source. In other words if a frontier point requires more relay points than robots available or if the frontier point cannot be traced back to the source then the frontier point is not appended to the ValidFrontierList. The algorithm also determines all neighbors of the candidate point based on WiFi distance threshold from the candidate point to each point in the FrontierList. The neighbors will be used to compute the possible paths.

The robot then checks to see if the ValidFrontierList is empty. If the list contains members then the robot chooses the first member as a candidate pose. The robot computes the possible paths from this candidate pose back to the source. Each path is calculated based on the neighbors of each frontier point in the FrontierList and contains a set of frontier points that create a path back to the WiFi source. The paths are calculated to ensure that only paths with poses less than or equal to the number of robots remaining in the hierarchy are considered. 

The robot chooses the candidate pose as it's chosen pose and finds the optimal path of this pose back to the source. The optimal path is calculated based on the heuristic
$$g(f) = \sum_i^k{(w_r(f_r^i) + w_n(f_n^i))}$$
where $g(f)$ is the cost associated with each path, $f_r^i$ is the rank value of each frontier point of the path determined by index value in the FrontierList, $f_n^i$ is a cost associated with including more frontier points in the path and is set to 1, and $w_r$ and $w_n$ are respective weights. The algorithm finds the optimal path based on two criteria; most prioritized frontier points and least dependency on other robots in the system.

Finally, if the robot fails to find a valid chosen frontier pose then it resorts to the backup choice algorithm presented in the next section.

\subsection{Backup Choice Algorithm}
The backup choice algorithm is designed to explore frontier points that otherwise do not contain a path that can be traced back to the WiFi source with pre-established frontier points. The algorithm chooses the first frontier point from the FrontierList and computes a shortest distance path from the pose to the WiFi source considering static obstacles (in the same way as a global planning algorithm). If the path distance is greater than the number of robots remaining multiplied by the WiFi distance threshold then the frontier point is not considered and the algorithm moves to the next frontier point. If the path distance is within the distance threshold then the algorithm computes poses for the minimum number of robots needed to relay the information back to the source. The algorithm provides a means to explore larger maps as it uses robots as a fireline to relay information from a distant pose back to the central computer. Finally, the robot may not find a viable frontier pose to choose, therefore the chosen pose defaults to the robot's starting position in the map and the pass down path becomes empty.

\begin{algorithm}
\caption{Backup Choice}
\begin{algorithmic}
\REQUIRE FrontierList
\STATE CandidateFrontierList := FrontierList
\WHILE{CandidateFrontierList not empty}
\STATE CandidatePose $\leftarrow$ DEQUEUE(CandidateFrontierList)
\STATE compute path from CandidatePose to WiFi source
\IF{path distance $<$ (robots remaining)*(WiFi range threshold)}
\STATE ChosenPose := CandidatePose
\STATE erase ChosenPose from FrontierList
\STATE find relay poses along path
\STATE append relay poses to PassDownPathList
\STATE return ChosenPose,FrontierList,PassDownPathList
\ENDIF
\ENDWHILE
\IF{fail to find ChosenPose}
\STATE ChosenPose $\leftarrow$ starting pose
\STATE PassDownPathList $\leftarrow \emptyset$
\STATE return ChosenPose, FrontierList, PassDownPathList
\ENDIF

\label{algorithm_choose_backup}
\end{algorithmic}
\end{algorithm}

\begin{figure}[H]
    \centering{
    \includegraphics[width=1.0\textwidth]{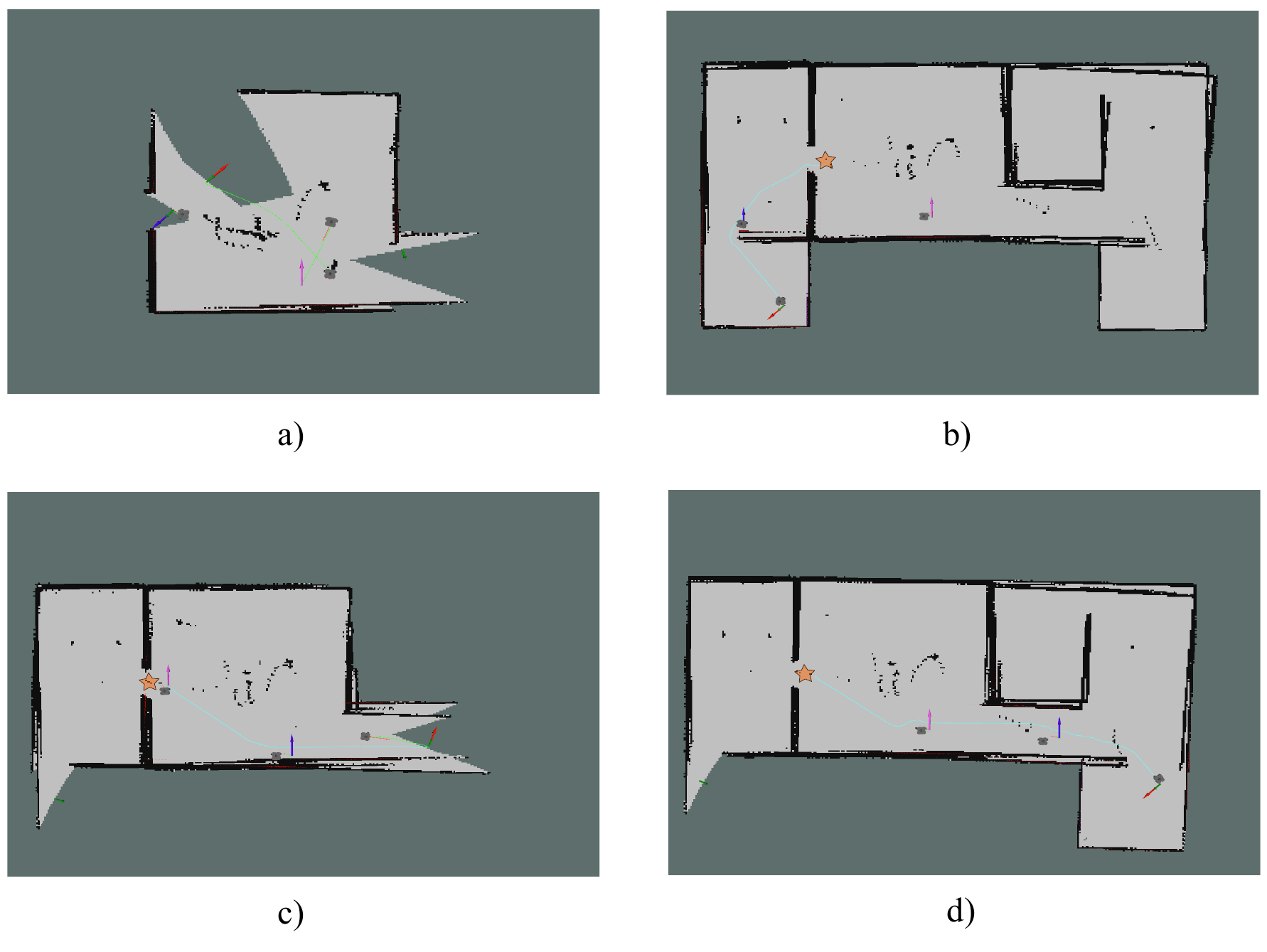}}
    \caption{A representation of the robot coordination algorithms. a) depicts the robots choosing frontier points based solely on the choose frontier point algorithm, b) shows a one robot relay using the backup choice algorithm, where robot 1 uses robot 2 as an information relay, c) and d) show a two robot relay where robot 1 uses robot 2 and 3 to relay information. The chosen poses are shown as red, blue, and pink arrows respectively and the WiFi source is denoted by the orange star}
    \label{fig:choice}
\end{figure}

\section{System Architecture}
\subsection{System Type}
The system type is based on the taxonomy of multi-robot systems shown in table \ref{tabel 1}. The first classification is a hybrid architecture. This system contains $n$ individual agents along with a central computer. Each of the agents generates local maps according to their own SLAM algorithms and sends the local map data to a central computer, which performs map merging, frontier point extrapolation, and sends frontier points to the individual agents for further exploration. 

The second classification is a homogeneous system. Each of the robots within this system has the exact same specifications. These include hardware and algorithms for performing SLAM, coordination, navigation, and communication.

The third classification is a cooperative system. Every robot works with every other robot to generate a map of the environment as efficiently as possible. This includes working together to optimally distribute tasks as well as prevent collisions during mobility. 

the fourth classification is a hierarchical information structure. The robots arrange themselves according to a hierarchy for decision making. The central computer computes optimal frontier points to be explored by the system. The central computer then passes the information to the prioritized robot, who makes a choice based on the data and then passes the unchosen points to the next robot on the priority list. This process repeats itself until all robots choose a task to perform.

The fifth classification is direct communication. All information sent from agent to agent occurs through on board hardware and a direct communication channel. Robots send all necessary data through an established communication protocol within the ROS framework either to the central computer or to one another.

The sixth and final classification is a reactive system architecture. Unforeseen changes in the environment are common especially as many robots navigate the environment all at once. In effect, each robot individually adjusts it's path plans according to unforeseen changes. This highlights the importance of a local planning algorithm for this type of architecture, because robots often need to adjust their course as other robots get in the way of their path.

\subsection{Hierarchy}

\begin{figure}[H]
    \centering{
    \includegraphics[width=1.0\textwidth]{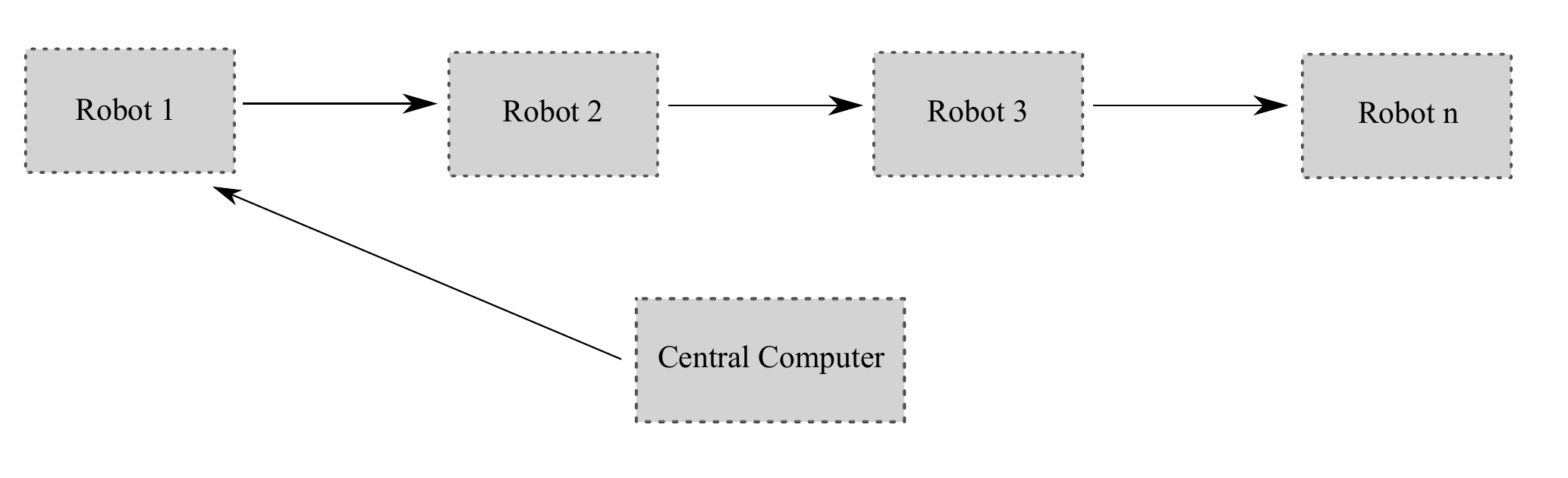}}
    \caption{An illustration of the hierarchical structure of this system. The central computer sends frontier points to the top robot in the hierarchy, which performs a choosing operation and sends all necessary information to the next robot.}
    \label{fig:hierarchy_diagram}
\end{figure}

The information structure of this system is a hierarchy. As seen in figure \ref{fig:hierarchy_diagram} the central computer sends a list of potential exploration (frontier) points to the first robot, which performs a necessary choosing operation and sends all pass down information to the next robot. The robots perform this procedure in sequence until all robots choose a task. 

Within the context of this project the hierarchy is organized arbitrarily due to a homogeneous system. There is no advantage to consider one robot as more capable than any other robot, because all robots contain the same specifications. However, in the case of heterogeneous robot systems the organization of this hierarchy becomes very important as it factors significantly into the performance.

There are a few advantages of using a hierarchy within this context. First, I assume that this architecture can be scaled to many agents. this means that a global optimal solution for robot exploration becomes an intractable problem for a large number of agents. A hierarchy allows the system to solve a series of local optimal solutions each within the ``nullspace" of the preceding solution. In this case the first robot chooses a frontier point that is optimal for itself. All less prioritized robots then choose local optimal frontier points while not interfering with the choice of preceding robots. The second advantage is that solutions can be computed with only local measurement data. Robots do not need to know the position of other robots when computing solutions. This frees up communication channels and prevents issues with incomplete state information.

The disadvantage of a hierarchy is a sacrifice of optimality for efficiency. The ``Get out of the way procedure" outlined above where each preceding robot makes a decision and all other robots only make decisions within the nullspace of established choices leaves out better team solutions. For example, robot 1 may choose a frontier point based on it's own optimal criteria and let all other robots choose based on the hierarchy, but it may be more beneficial for robot 1 to choose the frontier points for all other robots based on its own initial decision and state knowledge.

\subsection{Information Exchange}

\begin{figure}[H]
    \centering{
    \includegraphics[width=1.0\textwidth]{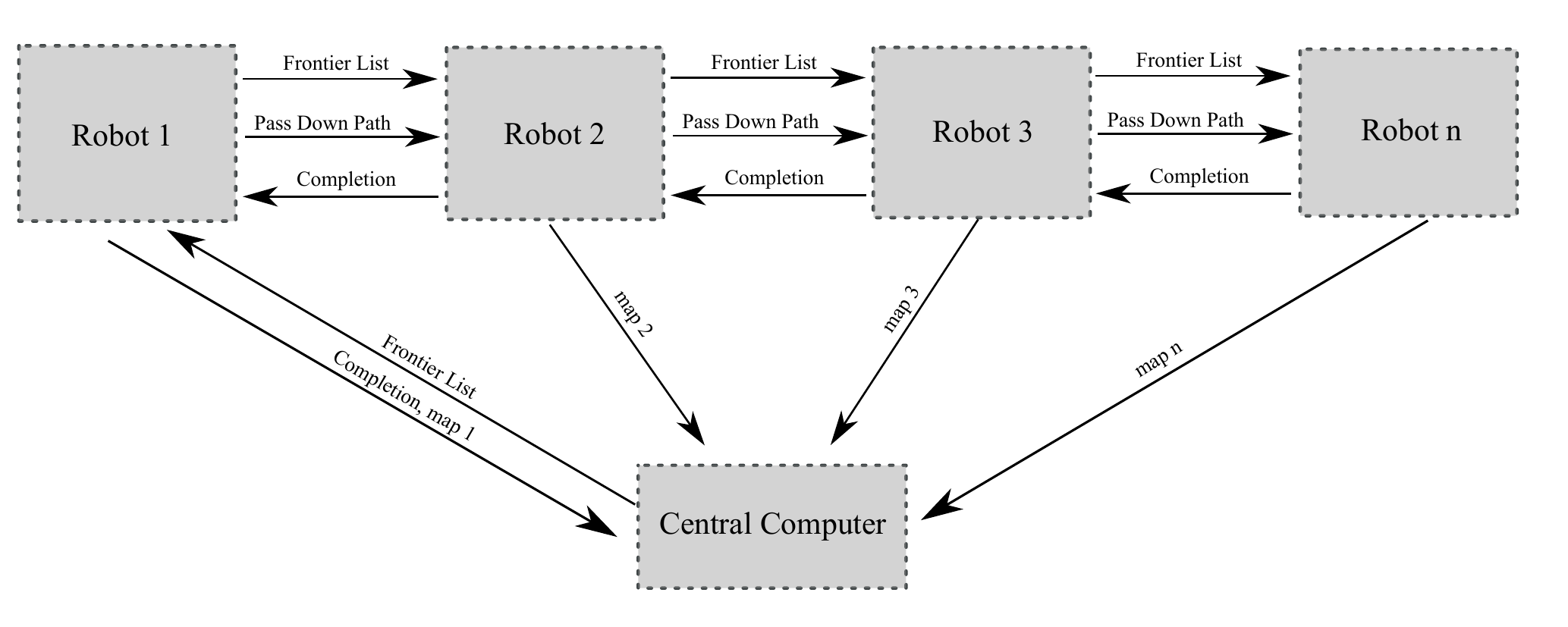}}
    \caption{An illustration of the information flow of this system. The central computer sends a list of frontier points to the first robot, which chooses a local optimal solution and sends a pass down path and the updated list of frontier points to the next robot. This process repeats until all robots choose a task. The robots send a completion notice back up the hierarchy after reaching the respective goal pose to indicate to the central computer to begin the next cycle}
    \label{fig:information_diagram}
\end{figure}

Information propagates through the hierarchy such that each robot chooses a local optimal task to achieve the global goal. 

The first question is what type of information is propagated in the system? The types of messages sent from the central computer to the robots or amongst the robots individually contain occupancy grid maps (each robot to the central computer), frontier point lists (central computer to robot and robot to robot), pass down path lists (robot to robot), and boolean completion messages (robot to robot and robot to central computer).

The second question is how does the information propagate? The central computer receives local map data from each of the robots individually in the form of occupancy grid map messages. The central computer computes optimal frontier points based on a merged map of all local maps received from the robots, and sends this list to the first robot in the hierarchy. After robot 1 receives the list of frontier points, it chooses a pose based on the algorithm in sections 3.3.2 and 3.3.3. Robot 1 then passes down two messages in the form of an updated frontier list and a pass down path (pass down path message will be empty if there is no robot dependency). This process is repeated until all robots choose a point. Finally, after all robots have completed their respective tasks a boolean completion message is sent back up the hierarchy until reaching the central computer as an indication to restart the process. The full information flow diagram is presented in \ref{fig:information_diagram}.

This information flow is advantageous because the most demanding information exchange occurs as occupancy grid map messages from each robot to the central computer. All other information exchange occurs as limited size arrays (frontier points, pass down paths) or booleans (completion messages).

\subsection{Communication Limits}
\begin{figure}[H]
    \centering{
    \includegraphics[width=1.0\textwidth]{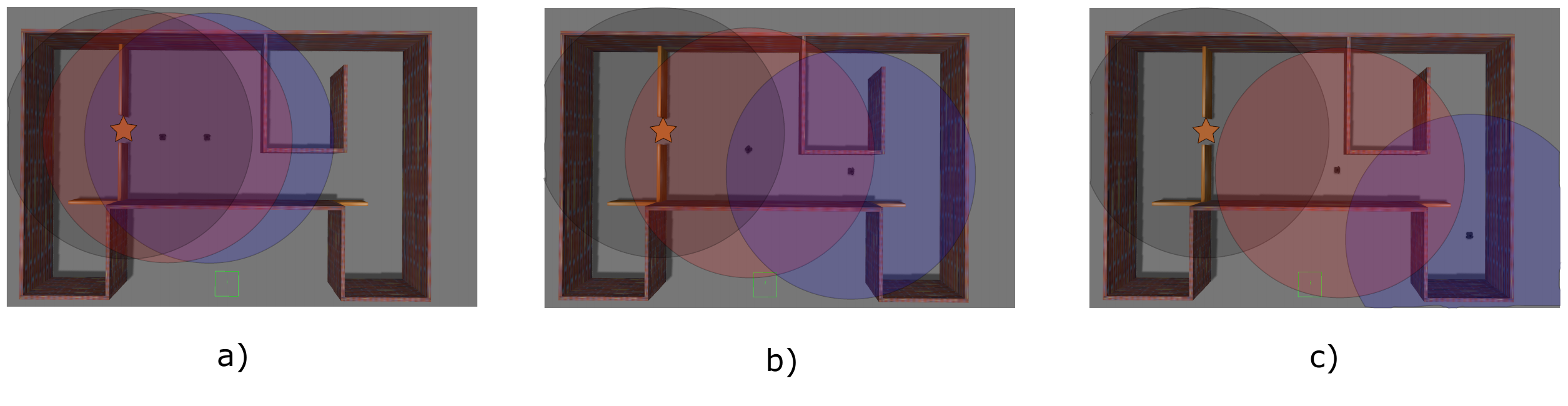}}
    \caption{A simulation of the WiFi range limitation. a) shows a viable structure where robot 1 (blue) and robot 2 (red) are within acceptable range of the WiFi source (star), b) shows a viable structure where robot 1 uses robot 2 as a relay for communication back to the source, and c) shows an invalid structure where robot 1 is not within acceptable range of robot 2 and robot 2 is not within acceptable range of the source}
    \label{fig:wifi_limit}
\end{figure}

There are two main communication constraints assumed by this system. First, each robot serves as a WiFi hotspot to send and receive data either to the central computer or to other robots. The WiFi range of each robot is limited, therefore robots can only communicate within a range of one another. Small indoor environments generally do not pose a problem, because WiFi has a range of about 100-150 feet for indoor environments, but as the size of space increases robots may need to explore areas that are beyond the available range of the on board WiFi module. This project aims to solve this issue by coordinating the robots as an efficient network relay structure. As the number of robots increases the available range of exploration also increases, because robots can relay information back to the central computer. A simulated example is shown in figure \ref{fig:wifi_limit}.

The second communication constraint is channel capacity of each robot. Although WiFi has a capacity of about 54Mbps and does not pose an issue for limited information exchange such as map data, limited channel capacity is a bottleneck for sending large data files such as video. In future work the second aim of this project is to exploit the network structure established by the robot coordination algorithm in order to provide the user with an option of increased channel capacity of all robots back to the source. The details are explained in the future work section.

\chapter{Results and Figures}

\section{Simulation Specifications}
\begin{figure}[H]
    \centering{
    \includegraphics[width=1.0\textwidth]{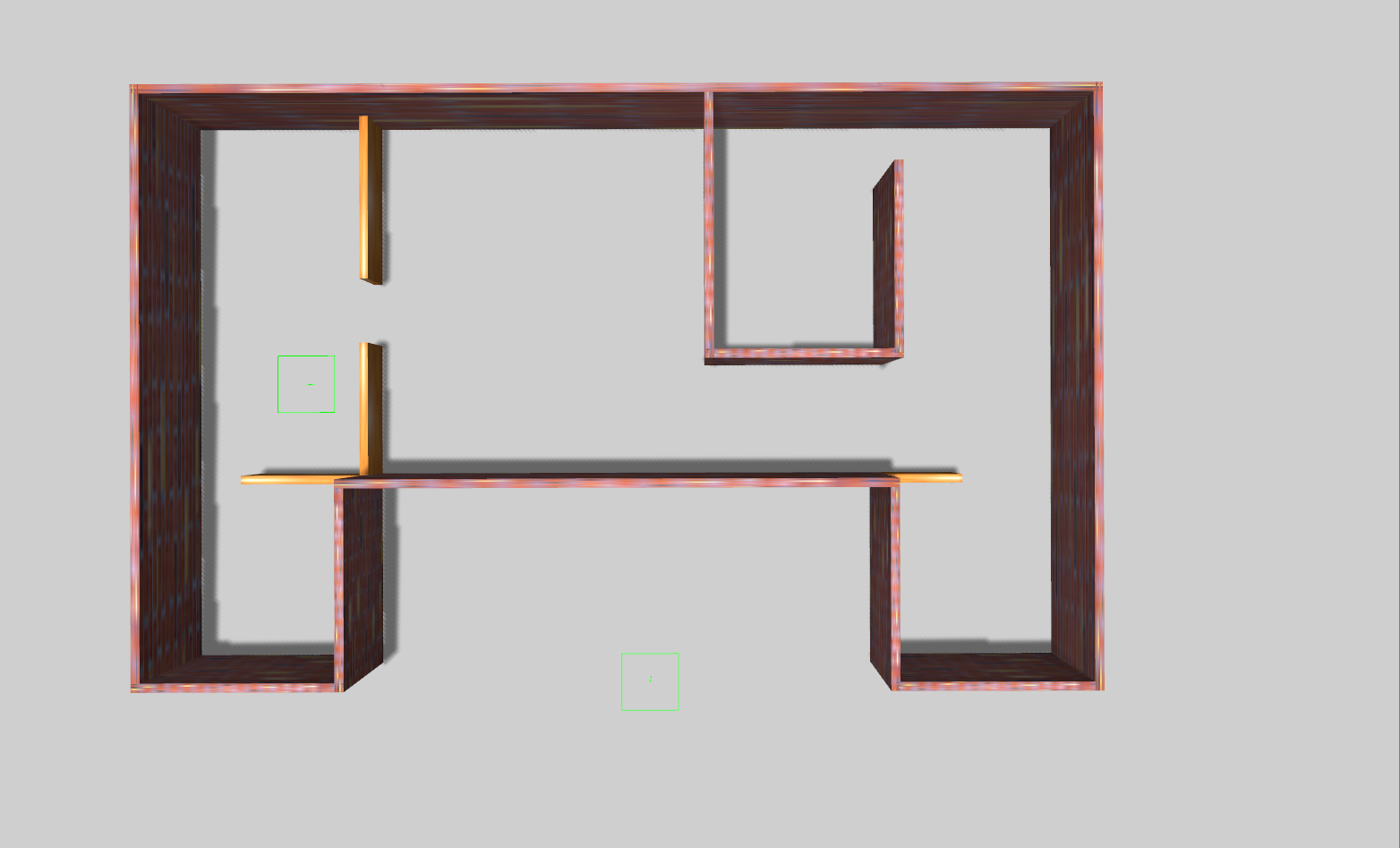}}
    \caption{The house map used for the first simulation}
    \label{fig:house_map}
\end{figure}

\begin{figure}[H]
    \centering{
    \includegraphics[width=1.0\textwidth]{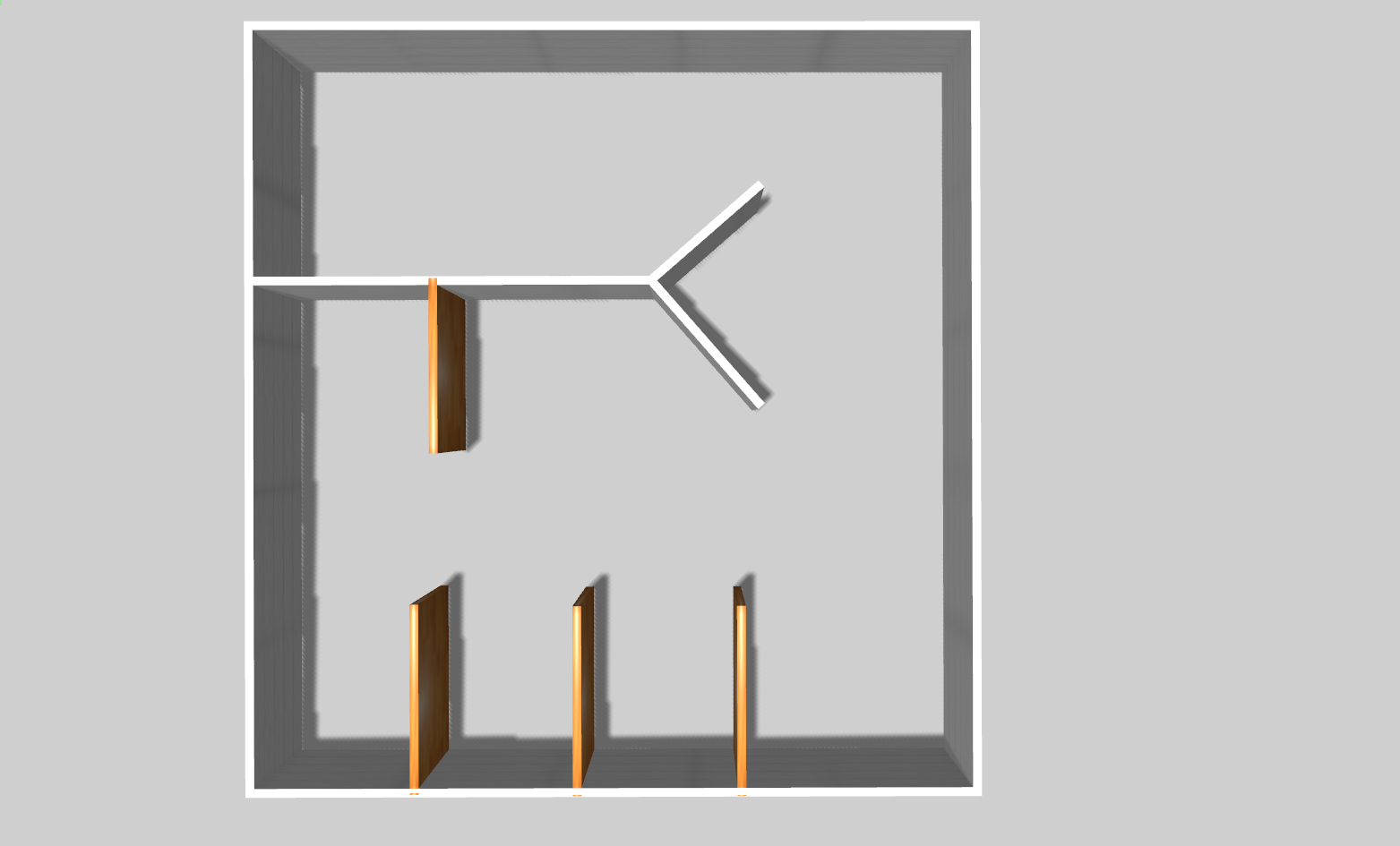}}
    \caption{The office map used for the second simulation}
    \label{fig:office_map}
\end{figure}

The simulation tests the performance of the architecture presented in chapter 3 in a ROS/Gazebo environment using two different maps; a house structure shown in \ref{fig:house_map} and an office environment shown in \ref{fig:office_map}. All simulated robots are based on the TurtleBot3 Waffle Pi from ROBOTIS in the Gazebo simulation environment. The simulation was performed using a desktop with an Intel Core i5 Processor and an NVIDIA GeForce RTX 2070 GPU running on a Linux Ubuntu operating system. All the code is written in C++.

The SLAM nodes are taken from the gmapping package from the ROS navigation stack. Each robot contains a gmapping node and all parameters remain consistent for all robots. In addition, the simulation uses the multirobot\_map\_merge package from the ROS navigation stack in order to perform map merging by the central computer. Finally, the robots use the move\_base package from the ROS navigation stack for navigation/path planning. 

The simulation runs for cases of 1,2,3, and 4 robots and varies the WiFi range parameters from 2m to 5m in 1m increments. The simulated WiFi range is not an accurate representation of actual WiFi range capabilities, but is diminished in order to test the capabilities of the algorithm for smaller maps. It is the assumption that the results here will remain at least fairly consistent with accurate real life WiFi range variations. Additionally, the simulation collects data for number of iterations of hierarchical choosing operations. An iteration is defined as the central computer extracting frontier points from the merged map, sending the list of frontier points down the robot hierarchy, each robot choosing a goal pose, and each robot completing their respective tasks. The simulation also notes the percentage of the map that the robot system completes based on each category. The map percentage completion is based on
$$ P = \frac{m_{partial}}{m_{complete}}*100$$
where $P$ is the percent completion, $m_{partial}$ is the number of known occupancy grid map values within the map completed by the robots, and $m_{complete}$ is the number of known occupancy grid map values for a completed map. The full table of results for each respective map is shown in tables \ref{table:results1} and \ref{table:results2}.

\newpage

\newpage
\section{Tables and Figures}

\begin{table}[H]
\caption{Simulation data for house map}
\centering
\begin{adjustbox}{width=1\textwidth}
\begin{tabular}{|l|l|l|l|}
\hline
Number of Robots & Number of Iterations & Map Completion Percentage & Simulated WiFi Range \\ \hline
1                & 5                    & 15.38                     & 2m                   \\
1                & 6                    & 21.32                     & 3m                   \\
1                & 10                   & 26.10                     & 4m                   \\
1                & 16                   & 58.21                     & 5m                   \\
2                & 10                   & 50.23                     & 2m                   \\
2                & 7                    & 51.12                     & 3m                   \\
2                & 14                   & 64.87                     & 4m                   \\
2                & 11                   & 69.44                     & 5m                   \\
3                & 7                    & 61.97                     & 2m                   \\
3                & 9                    & 71.90                     & 3m                   \\
3                & 7                    & 96.17                     & 4m                   \\
3                & 18                   & 100.00                    & 5m                   \\
4                & 10                   & 64.10                     & 2m                   \\
4                & 11                   & 90.68                     & 3m                   \\
4                & 15                   & 100.00                    & 4m                   \\
4                & 12                   & 100.00                    & 5m                   \\ \hline
\end{tabular}
\end{adjustbox}
\label{table:results1}
\end{table}

\begin{table}[H]
\caption{Simulation data for office map}
\centering
\begin{adjustbox}{width=1\textwidth}
\begin{tabular}{|l|l|l|l|}
\hline
Number of Robots & Number of Iterations & Map Completion Percentage & Simulated WiFi Range \\ \hline
1                & 4                    & 11.32                     & 2m                   \\
1                & 10                   & 16.94                     & 3m                   \\
1                & 14                   & 41.87                     & 4m                   \\
1                & 6                    & 30.67                     & 5m                   \\
2                & 8                    & 36.58                     & 2m                   \\
2                & 12                   & 50.59                     & 3m                   \\
2                & 16                   & 69.02                     & 4m                   \\
2                & 23                   & 73.81                     & 5m                   \\
3                & 14                   & 49.96                     & 2m                   \\
3                & 14                   & 71.45                     & 3m                   \\
3                & 29                   & 87.79                     & 4m                   \\
3                & 19                   & 100.00                    & 5m                   \\
4                & 12                   & 65.60                     & 2m                   \\
4                & 19                   & 100.00                    & 3m                   \\
4                & 27                   & 100.00                    & 4m                   \\
4                & 19                   & 100.00                    & 5m                   \\ \hline
\end{tabular}
\end{adjustbox}
\label{table:results2}
\end{table}

\begin{figure}[H]
    \centering{
    \includegraphics[width=1.0\textwidth]{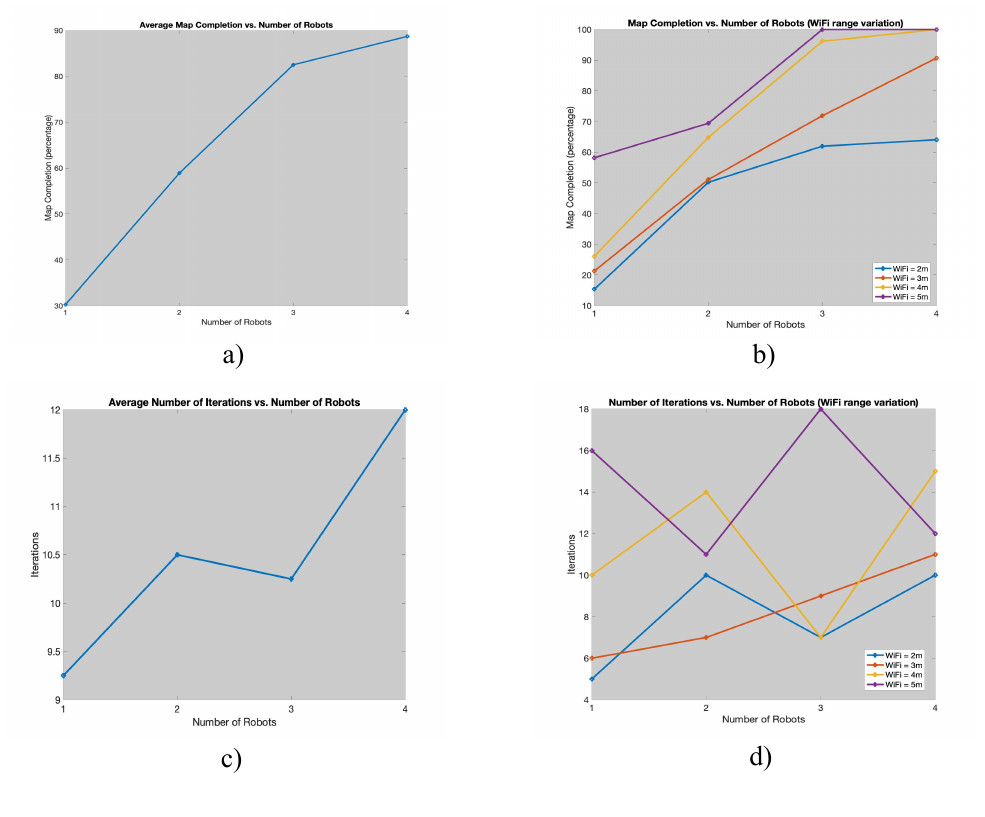}}
    \caption{Plots for the house map. a) shows the map completion percentage vs. number of robots averaged over the WiFi ranges, b) shows the map completion percentage vs. number of robots for each of the WiFi ranges, c) shows the number of iterations vs. number of robots averaged over all WiFi ranges, and d) shows the number of iterations vs. number of robots for each WiFi range}
    \label{fig:graphs1}
\end{figure}

\begin{figure}[H]
    \centering{
    \includegraphics[width=1.0\textwidth]{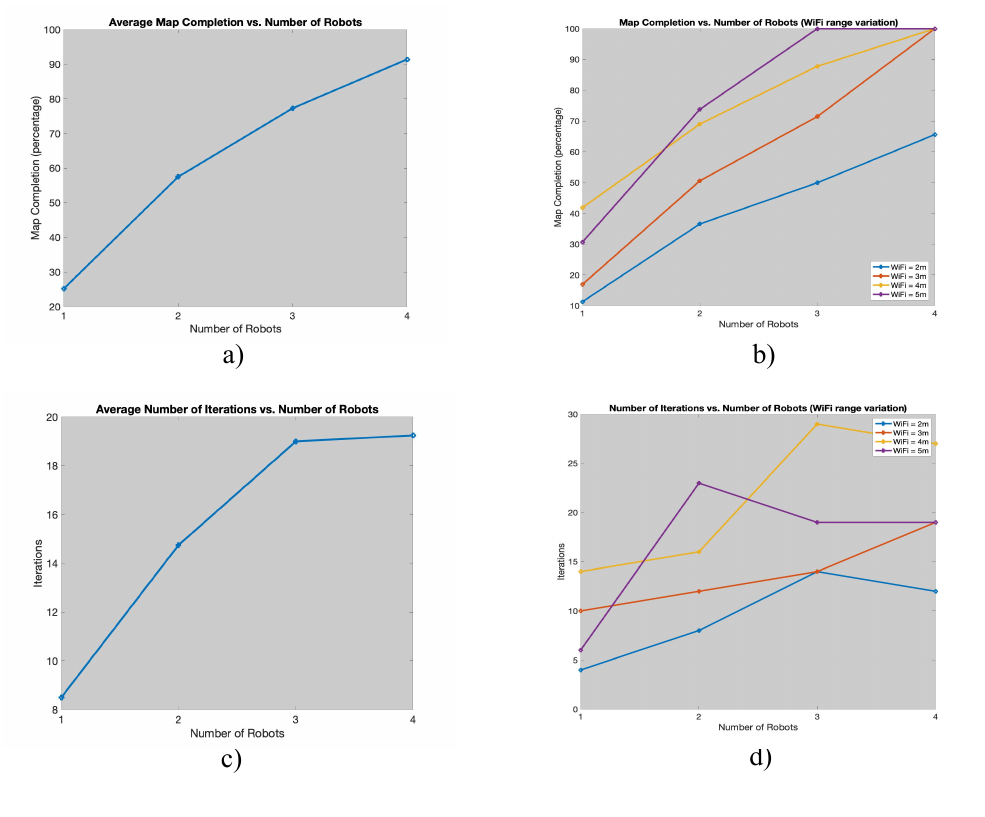}}
    \caption{Plots for the office map. a) shows the map completion percentage vs. number of robots averaged over the WiFi ranges, b) shows the map completion percentage vs. number of robots for each of the WiFi ranges, c) shows the number of iterations vs. number of robots averaged over all WiFi ranges, and d) shows the number of iterations vs. number of robots for each WiFi range}
    \label{fig:graphs2}
\end{figure}

\section{Discussion}
The performance between the two maps is fairly consistent. The map completion percentage shows a steady increase as the number of robots increases for all WiFi ranges based on figures \ref{fig:graphs1}b and \ref{fig:graphs2}b. Additionally, the average map completion over all WiFi ranges shows a steadily increasing behavior plateauting as the map completion percentage reaches 100 based on figures \ref{fig:graphs1}a and \ref{fig:graphs2}a. The question is whether the improved performance in map exploration outweighs the larger number of iterations with more robots in the system. The raw data shows that the number of iterations per simulation does not increase at a constant rate based on \ref{fig:graphs1}d and \ref{fig:graphs2}d, however the number of iterations as an average for the number of robots over WiFi range generally holds an upward trend based on \ref{fig:graphs1}c and \ref{fig:graphs2}c. As robots branch out to explore farther reaches of the map they rely heavily on the backup choice procedure, which in many instances only permits one robot to explore with incremental advances at each iteration. This is likely to account for the heavy increase in iterations as the number of iterations increase as more robots can branch out farther to explore the map.

Due to limited computing capacity a maximum of 4 robots were included in the simulation, which provides limited data for scalability of this architecture. In addition, as the number of robots increased the SLAM and map merging algorithms lost performance due to computational load. This may affect the outcome of the data as the 3 and 4 robot systems relied on a less accurate map. 

\chapter{Conclusion and Future Work}
The goal of this project is to develop a multi robot mapping architecture that takes into account communication limitations and adheres to three main criteria; efficiency, scalability, and robustness. 

The results of chapter 4 show that the performance significantly increases as the number of robots increase. The percentage of map completion consistently increases as more robots are added to the system and the number of coordination iterations is not a constant increase. In addition, the communication amongst robots is efficient. The robots send messages no larger than an occupancy grid map at each iteration. All other messages are small arrays and booleans. Finally, robots distribute tasks efficiently through the hierarchy. This includes information processing (SLAM algorithms, navigation, choice algorithms, frontier point extraction, and map merging) as well as frontier point allocations (choosing optimal frontier points to explore the map). 

The architecture can theoretically scale to large numbers of robots, however due to limited computation capacity only 4 robots were tested. Based on the results of chapter 4, the robots complete more of the mapping exploration as the number of agents increase and the communication burden does not significantly increase. This theoretically points to an ability to scale this architecture to many agents, but the data from this paper is not sufficient to prove this claim.

The system is robust to navigation challenges including unaccounted objects in the static map, but still has room to improve for communication errors, mapping errors, and mechanical/electrical failures. Each agent updates its navigation path based on unforeseen objects in the environment (such as other robots appearing in the calculated path) based on a local planning algorithm. This ensures that robots will consistently reach their chosen goal poses. However, the system does not adapt to local failures including communication delays/drops, mechanical/electrical failures, or mapping errors. In the future work section I describe how the architecture can be improved to account for dynamic and unpredictable challenges in the environment.

\subsubsection{Future Work}
I completed this project within a few months and unfortunately did not have the opportunity to implement some modules along the way. Here I will outline some of the next steps of this project.

The system did not adapt well to system failures including individual robot failures, communication drops, and software hiccups, so a next step is to implement coordinated backup features such that the system becomes less of a reactive architecture and more of a deliberative architecture. These backup features may reorganize the hierarchy if one robot fails, or coordinate the agents around communication failures.

In addition, the results and data collection are very limited. The next steps are to simulate the system with many more agents and extend the testing to real hardware. Real hardware can include both homogeneous and heterogeneous robots.

Because the algorithms developed in this paper are very general, I propose to extend the architecture to other high level search tasks. This may include multi-robot search and rescue, multi-robot person following, or other user defined tasks. 

Lastly, the communication amongst the robot network can be optimized. I have outlined some of the communication limitations with regard to channel capacity for large data sharing. Because this architecture arranges the robots in viable ``hotspot" relays for communication sharing at every iteration, it is possible to propagate information through the network using a max flow algorithm such as the ford-fulkerson algorithm. This may increase channel capacity for each robot and provide the user with the option to send large data files amongst the robots or back to the source. 

\addcontentsline {toc}{chapter}{Bibliography} 
\bibliography{bibliography}
\bibliographystyle{ieeetr}

\begin{thesisauthorvita}             
Henry Fielding Cappel was born on November 25, 1994 in Oak Park, Illinois. He attended school in the Chicago area through his highschool education and moved to Pennsylvania for his undergraduate education. He received a bachelor of arts from Swarthmore College located in Swarthmore, Pennsylvania in psychology with a minor in physics. After completing his bachelor's degree he decided that his true passion was in physics related to building robots.

Henry spent a year teaching robotics to children in the Chicago area travelling to various schools giving lessons for an after school robotics program. Although he loved to teach, his passion was to learn the most challenging robotics related problems and know how to build a complex robot from scratch. 

He decided to pursue his own degree from the University of Texas at Austin. He is receiving his masters of science in engineering in the field of aerospace engineering with a focus on robotics and control systems.
\end{thesisauthorvita}               

\end{document}

%% file: map_types.pdf_tex
\begingroup%
  \makeatletter%
  \providecommand\color[2][]{%
    \errmessage{(Inkscape) Color is used for the text in Inkscape, but the package 'color.sty' is not loaded}%
    \renewcommand\color[2][]{}%
  }%
  \providecommand\transparent[1]{%
    \errmessage{(Inkscape) Transparency is used (non-zero) for the text in Inkscape, but the package 'transparent.sty' is not loaded}%
    \renewcommand\transparent[1]{}%
  }%
  \providecommand\rotatebox[2]{#2}%
  \newcommand*\fsize{\dimexpr\f@size pt\relax}%
  \newcommand*\lineheight[1]{\fontsize{\fsize}{#1\fsize}\selectfont}%
  \ifx\svgwidth\undefined%
    \setlength{\unitlength}{626.45669291bp}%
    \ifx\svgscale\undefined%
      \relax%
    \else%
      \setlength{\unitlength}{\unitlength * \real{\svgscale}}%
    \fi%
  \else%
    \setlength{\unitlength}{\svgwidth}%
  \fi%
  \global\let\svgwidth\undefined%
  \global\let\svgscale\undefined%
  \makeatother%
  \begin{picture}(1,0.33031674)%
    \lineheight{1}%
    \setlength\tabcolsep{0pt}%
    \put(0,0){\includegraphics[width=\unitlength,page=1]{map_types.pdf}}%
    \put(0.07472554,0.04136304){\color[rgb]{0,0,0}\makebox(0,0)[lt]{\lineheight{1.25}\smash{\begin{tabular}[t]{l}a) Volumetric map\end{tabular}}}}%
    \put(0.40324869,0.03925747){\color[rgb]{0,0,0}\makebox(0,0)[lt]{\lineheight{1.25}\smash{\begin{tabular}[t]{l}b) Landmark map\end{tabular}}}}%
    \put(0.69775765,0.03925745){\color[rgb]{0,0,0}\makebox(0,0)[lt]{\lineheight{1.25}\smash{\begin{tabular}[t]{l}c) Occupancy grid map\end{tabular}}}}%
  \end{picture}%
\endgroup%

%% file: map_merge.pdf_tex
\begingroup%
  \makeatletter%
  \providecommand\color[2][]{%
    \errmessage{(Inkscape) Color is used for the text in Inkscape, but the package 'color.sty' is not loaded}%
    \renewcommand\color[2][]{}%
  }%
  \providecommand\transparent[1]{%
    \errmessage{(Inkscape) Transparency is used (non-zero) for the text in Inkscape, but the package 'transparent.sty' is not loaded}%
    \renewcommand\transparent[1]{}%
  }%
  \providecommand\rotatebox[2]{#2}%
  \newcommand*\fsize{\dimexpr\f@size pt\relax}%
  \newcommand*\lineheight[1]{\fontsize{\fsize}{#1\fsize}\selectfont}%
  \ifx\svgwidth\undefined%
    \setlength{\unitlength}{402.51968504bp}%
    \ifx\svgscale\undefined%
      \relax%
    \else%
      \setlength{\unitlength}{\unitlength * \real{\svgscale}}%
    \fi%
  \else%
    \setlength{\unitlength}{\svgwidth}%
  \fi%
  \global\let\svgwidth\undefined%
  \global\let\svgscale\undefined%
  \makeatother%
  \begin{picture}(1,0.97183099)%
    \lineheight{1}%
    \setlength\tabcolsep{0pt}%
    \put(0,0){\includegraphics[width=\unitlength,page=1]{map_merge.pdf}}%
    \put(0.34016888,0.02959007){\color[rgb]{0,0,0}\makebox(0,0)[lt]{\lineheight{1.25}\smash{\begin{tabular}[t]{l}d) Merged map\end{tabular}}}}%
    \put(0,0){\includegraphics[width=\unitlength,page=2]{map_merge.pdf}}%
    \put(0.0216719,0.68818672){\color[rgb]{0,0,0}\makebox(0,0)[lt]{\lineheight{1.25}\smash{\begin{tabular}[t]{l}a) Local map 1\end{tabular}}}}%
    \put(0.35329634,0.68813881){\color[rgb]{0,0,0}\makebox(0,0)[lt]{\lineheight{1.25}\smash{\begin{tabular}[t]{l}b) Local map 2\end{tabular}}}}%
    \put(0.69476133,0.69024666){\color[rgb]{0,0,0}\makebox(0,0)[lt]{\lineheight{1.25}\smash{\begin{tabular}[t]{l}c) Local map 3\end{tabular}}}}%
  \end{picture}%
\endgroup%

%% file: kinematicdiagram.pdf_tex
\begingroup%
  \makeatletter%
  \providecommand\color[2][]{%
    \errmessage{(Inkscape) Color is used for the text in Inkscape, but the package 'color.sty' is not loaded}%
    \renewcommand\color[2][]{}%
  }%
  \providecommand\transparent[1]{%
    \errmessage{(Inkscape) Transparency is used (non-zero) for the text in Inkscape, but the package 'transparent.sty' is not loaded}%
    \renewcommand\transparent[1]{}%
  }%
  \providecommand\rotatebox[2]{#2}%
  \newcommand*\fsize{\dimexpr\f@size pt\relax}%
  \newcommand*\lineheight[1]{\fontsize{\fsize}{#1\fsize}\selectfont}%
  \ifx\svgwidth\undefined%
    \setlength{\unitlength}{212.5984252bp}%
    \ifx\svgscale\undefined%
      \relax%
    \else%
      \setlength{\unitlength}{\unitlength * \real{\svgscale}}%
    \fi%
  \else%
    \setlength{\unitlength}{\svgwidth}%
  \fi%
  \global\let\svgwidth\undefined%
  \global\let\svgscale\undefined%
  \makeatother%
  \begin{picture}(1,0.62666667)%
    \lineheight{1}%
    \setlength\tabcolsep{0pt}%
    \put(0,0){\includegraphics[width=\unitlength,page=1]{kinematicdiagram.pdf}}%
    \put(0.12725198,0.10253967){\color[rgb]{0,0,0}\makebox(0,0)[lt]{\lineheight{1.25}\smash{\begin{tabular}[t]{l}$\theta_0$\end{tabular}}}}%
    \put(0.40136457,0.13639476){\color[rgb]{0,0,0}\makebox(0,0)[lt]{\lineheight{1.25}\smash{\begin{tabular}[t]{l}$\theta_1$\end{tabular}}}}%
    \put(0.53688558,0.2085659){\color[rgb]{0,0,0}\makebox(0,0)[lt]{\lineheight{1.25}\smash{\begin{tabular}[t]{l}$\theta_2$\end{tabular}}}}%
    \put(0.65095266,0.35093697){\color[rgb]{0,0,0}\makebox(0,0)[lt]{\lineheight{1.25}\smash{\begin{tabular}[t]{l}$\theta_3$\end{tabular}}}}%
    \put(0.05878997,0.57898259){\color[rgb]{0,0,0}\makebox(0,0)[lt]{\lineheight{1.25}\smash{\begin{tabular}[t]{l}y\end{tabular}}}}%
    \put(0.93532969,0.03972575){\color[rgb]{0,0,0}\makebox(0,0)[lt]{\lineheight{1.25}\smash{\begin{tabular}[t]{l}x\end{tabular}}}}%
    \put(0.43010147,0.07878332){\color[rgb]{0,0,0}\makebox(0,0)[lt]{\lineheight{1.25}\smash{\begin{tabular}[t]{l}h\end{tabular}}}}%
    \put(0.29277011,0.07626343){\color[rgb]{0,0,0}\makebox(0,0)[lt]{\lineheight{1.25}\smash{\begin{tabular}[t]{l}1\end{tabular}}}}%
    \put(0.45781967,0.18965627){\color[rgb]{0,0,0}\makebox(0,0)[lt]{\lineheight{1.25}\smash{\begin{tabular}[t]{l}2\end{tabular}}}}%
    \put(0.58803838,0.29995349){\color[rgb]{0,0,0}\makebox(0,0)[lt]{\lineheight{1.25}\smash{\begin{tabular}[t]{l}3\end{tabular}}}}%
    \put(0.67371302,0.4725626){\color[rgb]{0,0,0}\makebox(0,0)[lt]{\lineheight{1.25}\smash{\begin{tabular}[t]{l}4\end{tabular}}}}%
    \put(0.37907074,0.1749733){\color[rgb]{0,0,0}\makebox(0,0)[lt]{\lineheight{1.25}\smash{\begin{tabular}[t]{l}$\rho_1$\end{tabular}}}}%
    \put(0.5249291,0.25039823){\color[rgb]{0,0,0}\makebox(0,0)[lt]{\lineheight{1.25}\smash{\begin{tabular}[t]{l}$\rho_2$\end{tabular}}}}%
    \put(0.62320292,0.3826899){\color[rgb]{0,0,0}\makebox(0,0)[lt]{\lineheight{1.25}\smash{\begin{tabular}[t]{l}$\rho_3$\end{tabular}}}}%
  \end{picture}%
\endgroup%

%% file: task_hierarchy.pdf_tex
\begingroup%
  \makeatletter%
  \providecommand\color[2][]{%
    \errmessage{(Inkscape) Color is used for the text in Inkscape, but the package 'color.sty' is not loaded}%
    \renewcommand\color[2][]{}%
  }%
  \providecommand\transparent[1]{%
    \errmessage{(Inkscape) Transparency is used (non-zero) for the text in Inkscape, but the package 'transparent.sty' is not loaded}%
    \renewcommand\transparent[1]{}%
  }%
  \providecommand\rotatebox[2]{#2}%
  \newcommand*\fsize{\dimexpr\f@size pt\relax}%
  \newcommand*\lineheight[1]{\fontsize{\fsize}{#1\fsize}\selectfont}%
  \ifx\svgwidth\undefined%
    \setlength{\unitlength}{226.77165354bp}%
    \ifx\svgscale\undefined%
      \relax%
    \else%
      \setlength{\unitlength}{\unitlength * \real{\svgscale}}%
    \fi%
  \else%
    \setlength{\unitlength}{\svgwidth}%
  \fi%
  \global\let\svgwidth\undefined%
  \global\let\svgscale\undefined%
  \makeatother%
  \begin{picture}(1,0.375)%
    \lineheight{1}%
    \setlength\tabcolsep{0pt}%
    \put(0,0){\includegraphics[width=\unitlength,page=1]{task_hierarchy.pdf}}%
    \put(0.03189173,0.16933701){\color[rgb]{0,0,0}\makebox(0,0)[lt]{\lineheight{1.25}\smash{\begin{tabular}[t]{l}Task 1\end{tabular}}}}%
    \put(0.29768712,0.16945549){\color[rgb]{0,0,0}\makebox(0,0)[lt]{\lineheight{1.25}\smash{\begin{tabular}[t]{l}Task 2\end{tabular}}}}%
    \put(0.5718203,0.17043957){\color[rgb]{0,0,0}\makebox(0,0)[lt]{\lineheight{1.25}\smash{\begin{tabular}[t]{l}Task 3\end{tabular}}}}%
    \put(0.82929748,0.17184085){\color[rgb]{0,0,0}\makebox(0,0)[lt]{\lineheight{1.25}\smash{\begin{tabular}[t]{l}Task k\end{tabular}}}}%
  \end{picture}%
\endgroup%